\pdfoutput=1
\documentclass{article}
\usepackage{graphicx}
\usepackage{grffile}
\usepackage[final,nonatbib]{neurips_2021}

\usepackage[utf8]{inputenc} % allow utf-8 input
\usepackage[T1]{fontenc}    % use 8-bit T1 fonts
\usepackage{hyperref}       % hyperlinks
\usepackage{url}            % simple URL typesetting
\usepackage{booktabs}       % professional-quality tables
\usepackage{amsfonts}       % blackboard math symbols
\usepackage{nicefrac}       % compact symbols for 1/2, etc.
\usepackage{microtype}      % microtypography
\usepackage{xcolor}         % colors
\usepackage{colortbl}

\usepackage{enumitem}
\usepackage[small]{caption}
\usepackage{subcaption}

\usepackage{multirow}
\usepackage{appendix}
\usepackage{amsmath,amssymb,amsthm,mathtools}
\usepackage{wrapfig}
\usepackage{algorithm, algorithmic}
\usepackage{pifont}

\usepackage{amsmath,amsfonts,bm}

\def\eqref#1{equation~\ref{#1}}
\def\1{\bm{1}}

\DeclareMathAlphabet{\mathsfit}{\encodingdefault}{\sfdefault}{m}{sl}
\SetMathAlphabet{\mathsfit}{bold}{\encodingdefault}{\sfdefault}{bx}{n}

\def\sI{{\mathbb{I}}}

\newcommand{\hatpi}{\hat{\pi}}
\newcommand{\hatrhob}{\rho_{\hat{\pi}_b}}
\newcommand{\E}{\mathbb{E}}

\newcommand{\TV}{D_{\mathrm{TV}}}

\newtheorem{definition}{Definition}%[section]
\newtheorem{lemma}{Lemma}%[section]
\newtheorem{theorem}{Theorem}%[section]
\newtheorem{proposition}{Proposition}%[section]
\newtheorem{observation}{Observation}

\newcommand{\fig}[1]{Fig.~\ref{#1}}
\newcommand{\eq}[1]{Eq.~(\ref{#1})}
\newcommand{\ineq}[1]{Ineq.~(\ref{#1})}
\newcommand{\tb}[1]{Tab.~\ref{#1}}
\newcommand{\se}[1]{Section~\ref{#1}}
\newcommand{\ap}[1]{Appendix~\ref{#1}}

\newcommand{\lm}[1]{Lemma~\ref{#1}}
\newcommand{\prop}[1]{Proposition~\ref{#1}}
\newcommand{\alg}[1]{Algo.~\ref{#1}}
\newcommand{\theo}[1]{Theorem~\ref{#1}}

\newcommand{\observe}[1]{Observation~\ref{#1}}

\newcommand{\bbI}{\ensuremath{\mathbb{I}}} % Indicator
\newcommand{\bbE}{\ensuremath{\mathbb{E}}} % Expect
\newcommand{\caA}{\ensuremath{\mathcal{A}}} % Action
\newcommand{\caS}{\ensuremath{\mathcal{S}}} % State
\newcommand{\caM}{\ensuremath{\mathcal{M}}} % Model
\newcommand{\caD}{\ensuremath{\mathcal{D}}} 
\newcommand{\caL}{\ensuremath{\mathcal{L}}} 
\newcommand{\caT}{\ensuremath{\mathcal{T}}} 
\newcommand{\caB}{\ensuremath{\mathcal{B}}} 
\newcommand{\kld}{\text{D}_{\text{KL}}}

\newcommand{\hytt}[1]{\texttt{\hyphenchar\font=\defaulthyphenchar #1}}

\title{Curriculum Offline Imitating Learning}

\author{%
   Minghuan Liu$^{1}$\thanks{Equal contribution. \dag Corresponding author. Codes are available at \url{https://github.com/apexrl/COIL}.}~~~ Hanye Zhao$^{1*}$~~~ Zhengyu Yang$^{1}$~~~ Jian Shen$^{1}$~~~\\
   \textbf{Weinan Zhang$^{1\dag}$~~~ Li Zhao$^{2}$~~~ Tie-Yan Liu$^{2}$}\\
  {$^1$ Shanghai Jiao Tong University, $^2$ Microsoft Research}\\
  \texttt{\{minghuanliu, fineartz, zyyang, rockyshen, wnzhang\}@sjtu.edu.cn,} \\
  \texttt{\{lizo,tyliu\}@microsoft.com}
 }

\begin{document}

\maketitle

\begin{abstract}

Offline reinforcement learning (RL) tasks require the agent to learn from a pre-collected dataset with no further interactions with the environment. 
Despite the potential to surpass the behavioral policies, RL-based methods are generally impractical due to the training instability and bootstrapping the extrapolation errors, which always require careful hyperparameter tuning via online evaluation. 
In contrast, offline imitation learning (IL) has no such issues since it learns the policy directly without estimating the value function by bootstrapping. 
However, IL is usually limited in the capability of the behavioral policy and tends to learn a mediocre behavior from the dataset collected by the mixture of policies. 
In this paper, we aim to take advantage of IL but mitigate such a drawback. 
Observing that behavior cloning is able to imitate neighboring policies with less data, we propose \textit{Curriculum Offline Imitation Learning (COIL)}, which utilizes an experience picking strategy for imitating from adaptive neighboring policies with a higher return, and improves the current policy along curriculum stages. 
On continuous control benchmarks, we compare COIL against both imitation-based and RL-based methods, showing that it not only avoids just learning a mediocre behavior on mixed datasets but is also even competitive with state-of-the-art offline RL methods.
\end{abstract}

\section{Introduction}
\label{sec:intro}

Offline reinforcement learning (RL), or batch RL, aims to learn a well-behaved policy from arbitrary datasets without interacting with the environment. This setting is generally a more practical paradigm than online RL since it is expensive or dangerous to interact with the environment in most real-world applications. Typically, two main kinds of offline datasets are considered in previous offline RL works~\cite{fu2020d4rl,qin2021neorl}: one contains transitions sampled by a single behavioral policy; the other includes a buffer collected by a mixture of policies. 
% \li{Trajectorys are different from a buffer of data, the later could be transition samples without trajectory information.  One ... the other ... }

Two main approaches have been deeply investigated for Offline RL. 
% \li{Should we introduce model-based methods for offline RL? Is the imitation learning based method deeply investigated? I did not see imitation learning based methods in Offline RL tutorial. }
First, RL-based methods, in particular, Q-learning and policy gradient-based algorithms~\cite{kumar2020conservative,fujimoto2019off,kumar2019stabilizing}, have the potential to outperform the behavioral policy. 
% \li{Is this claim correct? I think recent work on Offline RL only prove that with pessimistic method we can find the optimal policy in case the optimal policy is in the set of behavior policies. I am not sure if learning with offline RL can actually achieve better performance than the best performance in offline data. }
However, they always suffer from serious bootstrapping errors and training instability. This shortcoming makes such algorithms impractical to be utilized since too many hyperparameters need to be tuned to achieve a good performance, and it is hard to evaluate a suitable model in an offline manner, as revealed in \cite{qin2021neorl}.
In contrast, offline imitation learning~\cite{bain1995framework,peng2019advantage,chen2020bail}, specifically, behavior cloning (BC), can always stably learn to perform as the behavioral policy, which may be helpful under single-behavior datasets. However, BC may fail in learning a good behavior under a diverse dataset containing a mixture of policies (both goods and bads).

\textbf{Quantity-quality dilemma on mixed dataset.} 
% \li{Is this dilemma well-known? Or is it proposed by us? If proposed by us, we should introduce the dilemma. }
As a supervised learning technique, BC is not easy to fulfill a desired result, especially on a mixed dataset. Specifically, it requires both quantity and quality of the demonstration data, which can hardly be satisfied in offline RL tasks. Directly mimicking the policy from a mixed dataset that contains bad-to-good demonstrations can be regarded as imitation learning from a mediocre behavior policy. To achieve the best performance of the dataset, a naive idea is to imitate the top trajectories ordered by its return. However, such a simple strategy will reach the \textit{quantity-quality dilemma} on the mixed dataset. For example, Fig. 1a illustrates the ordered trajectories on the \texttt{Walker-final-dataset}, which contains the whole training experience sampled by an online training agent. Different BC agents are trained by the top 10\%, 25\%, 50\%, and 100\% trajectories, but none of them gets rid of a mediocre performance, as shown in Fig. 1b. Typically, on such dataset, less data owns higher quality but less quantity, and thus cause serious compounding error problems~\cite{ross2011reduction,ho2016generative,fu2018learning}; on the other hand, more data provides a larger quantity, yet its mean quality becomes worse. 
In this work, we aim to solve such a dilemma and exploit the most potential of IL to derive a stable and practical algorithm reaching the best performance of a given dataset.

Our intuition comes from the observation that under RL scenarios, the agent can imitate a neighboring policy with much fewer samples. %, as a motivating example illustrated in \se{sec:motivation} 
This observation promotes a curriculum solution for the above challenge. Specifically, for mixed datasets, the agent can adaptively imitate the better neighboring policies step by step and finally reach the optimal behavior policy of the dataset.

% Our intuition comes from the observation of an online agent training, where the agent begins with a poorly behaved policy and improves it gradually guided by the reward function. Each training iteration can be regarded as imitating an optimal policy defined by the reward function. In offline setting, where we have pre-collected data by various quality of policies, we can let the agent adaptively imitate the neighboring better policies to make full use of the experience and finally get reach to the best policy.

\begin{wrapfigure}{r}{0.6\textwidth}
\vspace{-13pt}
\begin{minipage}[b]{0.49\linewidth}
\centering
{\includegraphics[height=0.9\linewidth]{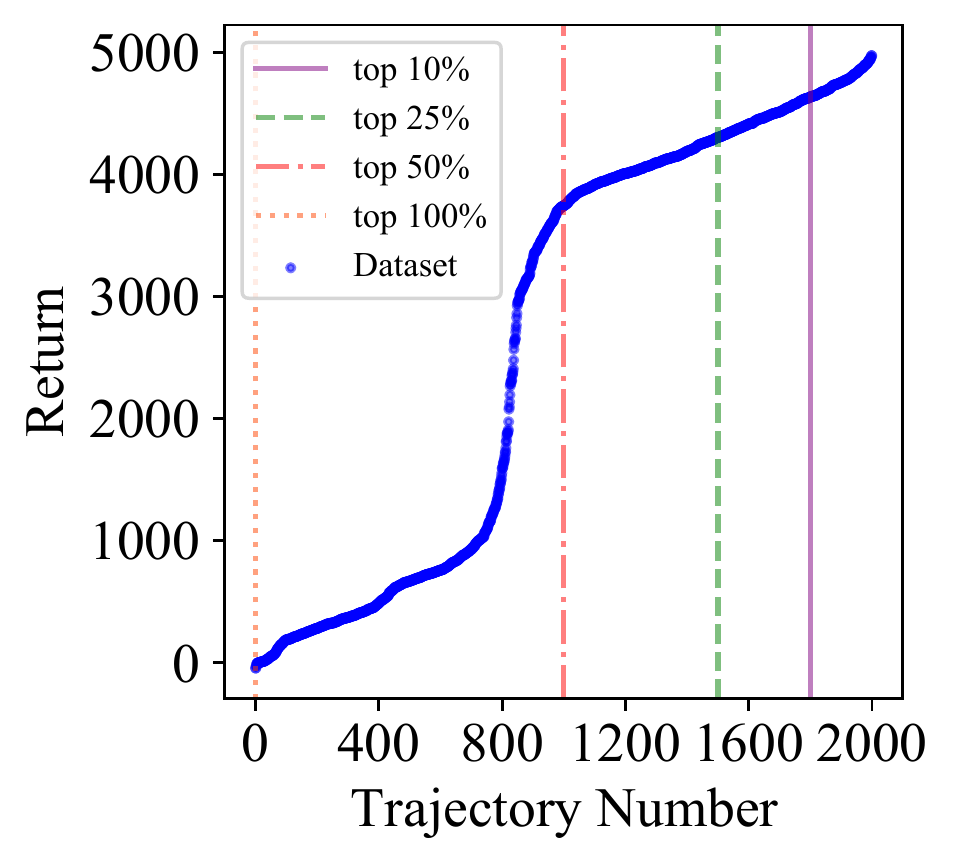}

\vspace{-7pt}
\caption*{(a) Ordered trajectories.}
\label{fig:example-suba}
}
\end{minipage}
\begin{minipage}[b]{0.49\linewidth}
\centering
{\includegraphics[height=0.9\linewidth]{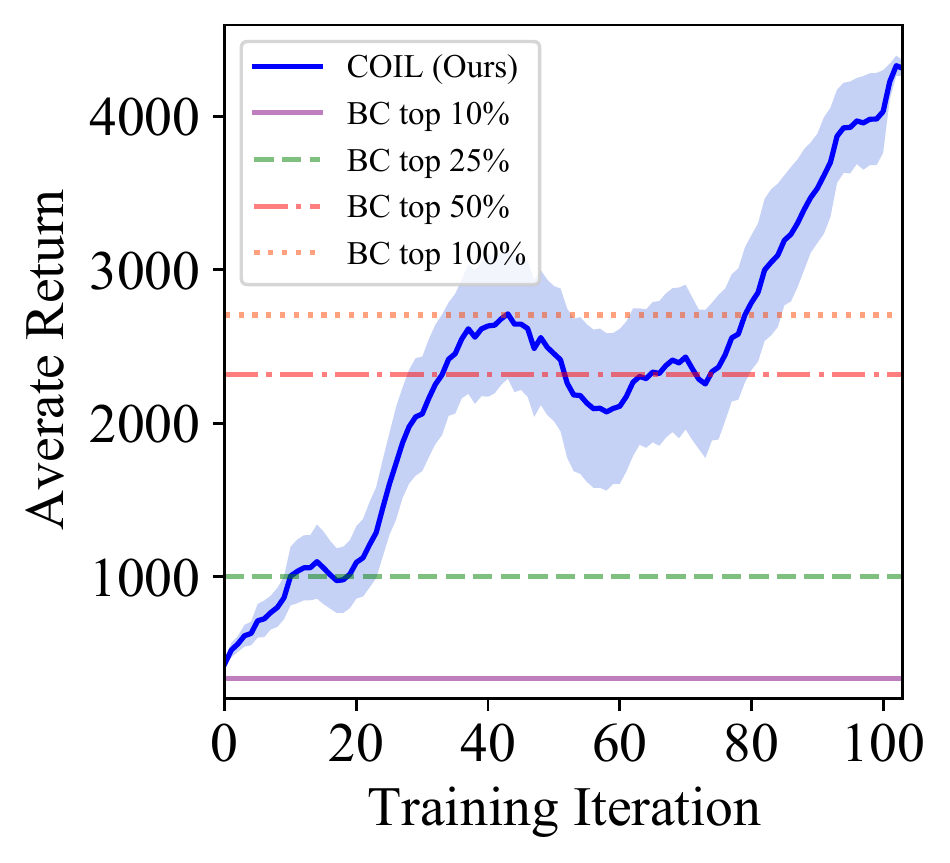}

\vspace{-7pt} 
\caption*{(b) Returns of BC v.s. COIL.}
\label{fig:example-subb}
}
\end{minipage}
\vspace{-6pt}
\caption{Examples of the quality-quantity dilemma for BC. (a) Trajectories of the Walker2d-final dataset ordered by their accumulated return. (b) Performances of behavior cloning (BC) for learning the top 10\%, 25\%, 50\%, and 100\% trajectories of the dataset.}
\vspace{-10pt}
\label{fig:example}
\end{wrapfigure}

\textbf{Our work.} We propose Curriculum Offline Imitation Learning (COIL), a simple yet effective offline imitation learning framework for offline RL. At each training iteration, COIL improves the current policy with the data sampled by neighboring policies. To achieve that, COIL utilizes an adaptive experience picking strategy and a return filter to select proper trajectories from the offline dataset for the current level of the agent and thus produces stages of the curriculum. Notably, COIL stops with a close-to-data-optimal policy without finding the best model under online evaluation. This feature allows to deploy the algorithm in practical problems.
In experiments, we show the effectiveness of COIL on various kinds of offline datasets. Fig. 1 offers a quick review of our results: depending solely on BC, COIL can learn from scratch to reach the best performance of the given dataset.

\textbf{Contributions.} To summarize, the main technical contributions of this paper are as follows.
\begin{itemize} [leftmargin=0.2in]
\vspace{-5pt}
        \item We highlight how the discrepancy between the target policy and the initialized policy affects the number of samples required by BC (\se{sec:motivation});
    \vspace{-3pt}

    \item Depending on BC, we propose a practical and effective offline RL algorithm with a practical neighboring experience picking strategy that ends with a good policy (\se{sec:method});
    \vspace{-3pt}
    
    \item We present promising comparison results with comprehensive analysis for our algorithm, which is competitive to the state-of-the-art methods (\se{sec:experiment}).
\end{itemize}

\section{Preliminaries}
\label{sec:pre}

\paragraph{Notations.}

We consider a standard Markov Decision Process (MDP) as a tuple $\caM = \langle \caS, \caA, \caT, \rho_0, r, \gamma \rangle$, where $\caS$ is the state space, $\caA$ represents the action space, $\caT: \caS \times \caA \times \caS \rightarrow [0, 1]$ is the state transition probability distribution, $\rho_0: \caS\rightarrow[0,1]$ is the initial state distribution, $\gamma\in [0,1]$ is the discounted factor, and $r: \caS \times \caA \rightarrow \mathbb{R}$ is the reward function. The goal of RL is to find a policy $\pi(a|s): \caS \times \caA \rightarrow [0, 1]$ that maximizes the expected cumulative discounted rewards (or called return) along a trajectory $\tau$: $R(\tau)=\sum_{t=0}^{T}\gamma^t r_t$. 
The dataset $\caD$ consists of trajectories $\{\tau_{1}^N\}$ that are the collected by a mixture of bad-to-good policies, where a trajectory $\tau_i = \{(s^{i}_{0}, a^{i}_{0}, s'^{i}_{0}, r^{i}_{0}), (s^{i}_{1}, a^{i}_{1}, s'^{i}_{1}, r^{i}_{1}), \cdots, (s^{i}_{h_i}, a^{i}_{h_i}, s'^{i}_{h_i}, r^{i}_{h_i})\}$, and $h_i$ is the horizon of $\tau_i$. For any dataset, we assume a behavior policy $\pi_b$ that collects such data and its empirical estimation $\hatpi_b$ can be induced from $\caD$ as $\hatpi_b(a|s)=\frac{\sum_{(s',a')\in\caD}\sI[s'=s,a'=a]}{\sum_{s'\in\caD}\sI[s'=s]}$, where $\sI$ is the indicator function.
We further introduce a common used term, occupancy measure, which is defined as the discounted occurrence probability of states or state-action pairs under policy $\pi$: 
$\rho_{\pi}(s,a) = \sum_{t=0}^{\infty}\gamma^t P(s_t=s, a_t=a|\pi) = \pi(a|s)\sum_{t=0}^{\infty}\gamma^t P(s_t=s|\pi)
= \pi(a|s)\rho_{\pi}(s)$. With such a definition we can write down that $\bbE_{\pi}[\boldsymbol{\cdot}]=\sum_{s,a}\rho_{\pi}(s,a)[\boldsymbol{\cdot} ]=\bbE_{(s,a)\sim \rho_{\pi}}[\boldsymbol{\cdot}]$.

\begin{definition}
The partial order of different policies is defined as the relative return quantity that a policy can achieve when deployed in the environment. Formally, given two policies $\pi_1$ and $\pi_2$:
\begin{align}
    \pi_1 \preceq \pi_2 \Leftrightarrow  R_1 \preceq R_2
\end{align}
\end{definition}
Therefore, by definition, in a mixed dataset $\caD$ that is collected by $K$ different policies $\pi_1, \cdots, \pi_K$, the optimal behavior policy $\pi^*$ can be determined such that for $\forall i\in[1,K], \pi_i \preceq \pi^*$.

\paragraph{Curriculum learning.}
Curriculum learning design and construct a \emph{curriculum} automatically as a sequence of tasks $G_1, \dots, G_N$ to train on, such that the efficiency or performance on a target task $G_t$ is improved. The expected loss on the $j^{\text{th}}$ task is denoted $\caL_j$.

% \begin{definition}
% The distance between policies are defined by the statistical distance of the distribution of their sampled trajectories. Formally, given policies $\pi_1$ and $\pi_2$:
% \begin{align}
%     d(\pi_1, \pi_2) \triangleq  d(P_{\pi_1}(\tau), P_{\pi_2}(\tau))
% \end{align}
% \end{definition}

% \paragraph{Occupancy Measure.}
% In RL literature, researchers usually utilize the  concept of occupancy measure (OM)~\cite{ho2016generative} to characterize the statistical properties of an MDP under a certain policy. Specifically, the state OM and the state-action OM are defined as the density over the states and state-action pairs under a given policy $\pi$:
% \begin{equation}
% \begin{aligned}
%     \rho_{\pi}(s) &= \sum_{t=0}^{\infty}\gamma^t P(s_t=s|\pi)\\
%      \rho_{\pi}(s,a) &= \sum_{t=0}^{\infty}\gamma^t P(s_t=s,a_t=a|\pi)=\rho_{\pi}(s)\pi(a|s)~.
% \end{aligned}
% \end{equation}

\section{Empirical Observations and Theoretic Analysis}
\label{sec:motivation}

In this section, we begin with empirical observations that motivate the core idea of our method, followed by the theoretical analysis to support our motivation. 
Generally, we aim to investigate how the asymptotic performance of BC is affected by the discrepancy between the demonstrated policy and the initialized imitating policy. Previous research shows that BC tends to fail with a small number of high-quality demonstrations but can learn well from large-quantity and high-quality data~\cite{ho2016generative,ghasemipour2020divergence}. On the contrary, we find that the requirement of quantity can be highly relaxed as the similarities between the demonstrated policy and the initialized imitating policy increase. 

\begin{figure}[htbp]
\vspace{-7pt}

\begin{subfigure}[b]{0.33\textwidth}
\centering
\includegraphics[height=0.85\linewidth]{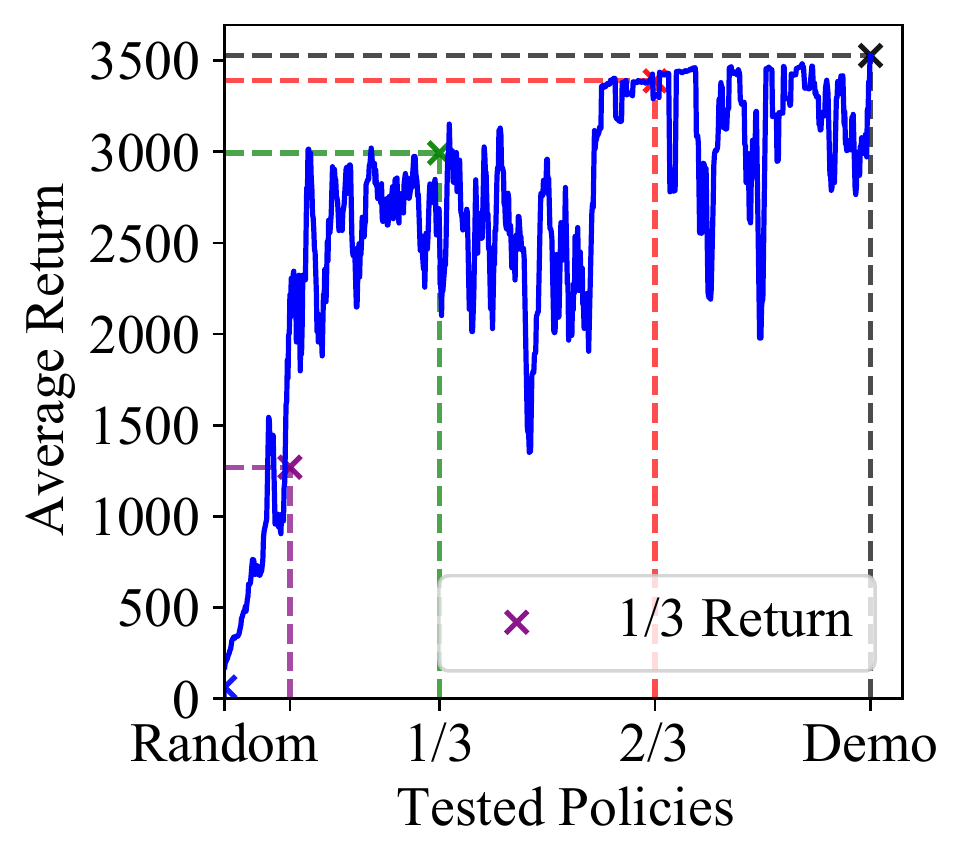}

\vspace{-5pt}
\caption{Online training curve.}
\label{fig:sac-training}
\end{subfigure}
\begin{subfigure}[b]{0.33\textwidth}
\centering
\includegraphics[height=0.85\linewidth]{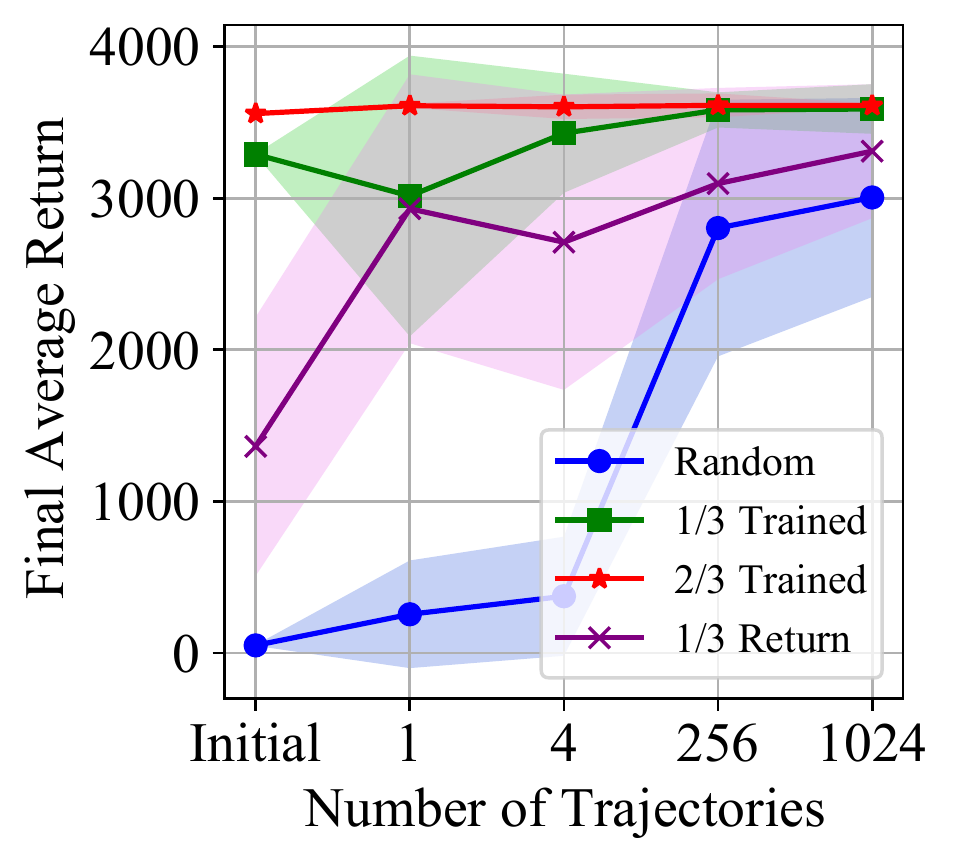}

\vspace{-5pt} 
\caption{Final performance of BC.}
\label{fig:motivation-suba}
\end{subfigure}
\begin{subfigure}[b]{0.33\textwidth}
\centering
\includegraphics[height=0.85\linewidth]{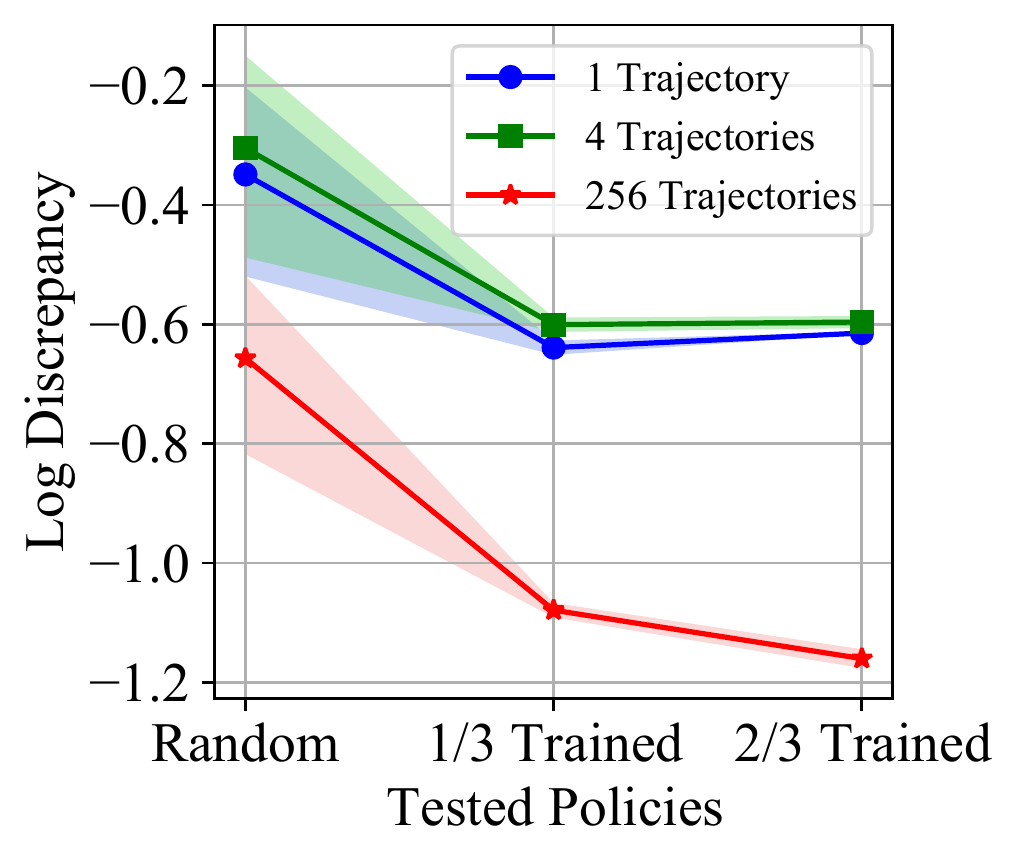}

\vspace{-5pt} 
\caption{Empirical discrepancy.}
\label{fig:motivation-subb}
\end{subfigure}

\vspace{-5pt}
\caption{(a) Online training curves of an SAC agent trained on the Hopper environment, where the crosses and dashed lines indicate the stage of selected policies. (b) Final performances achieved by imitating the demo policy using BC, initialized with different stages of policies. The curves depict the fact that close-to-demonstration policy can easily imitate the demonstrated policy with fewer samples. (c) Empirical estimation on the discrepancy between the initialized policy and the trained policy outside the support of the demonstrations. Initialized with a closer-to-demo policy always enjoys more minor discrepancy. }
\vspace{-7pt}
\label{fig:motivation}
\end{figure}

\subsection{BC with Different Initialization}\label{subsec:bc}

To construct the motivating example, we choose Hopper as the testbed, and train an SAC agent until convergence to sample various counts of trajectories as the demonstration data. We then take the online-trained policy checkpoints at different training iterations as the initiated policy to train an IL agent. Particularly, the first agent adopts the \texttt{Random} policy to imitate the demonstration by BC; the second uses the policy of \texttt{1/3 Return} and the other two agents start with \texttt{1/3 Trained} and \texttt{2/3 Trained} policy separately, in terms of training iterations (see \fig{fig:sac-training}). 

The results are shown in \fig{fig:motivation-suba}, where we illustrate the average return of each agent given the different number of demonstrations (the exact quantitative results can be found in \ap{ap:motivation-results}). The results show that initialized with a \texttt{Random} policy, the agent can only learn to imitate the demonstrated policy well with a large number of samples\footnote{We note that the \texttt{Random} agent can only work well on large datasets with normalized state space; however the other agents learn well upon raw states.}; in addition, both \texttt{1/3 Return} and \texttt{1/3 Trained} policies can achieve a sub-optimal performance with fewer samples, where the \texttt{1/3 Trained} one is more efficient. In comparison, the \texttt{2/3 Trained} agent, which is the closest to the optimal policy, can stably achieve the best performance with all amounts of trajectories.

\subsection{Theoretical Analysis}
Beyond these observations, we are inquisitive to find a theoretical explanation to support our claim. Standing on the primary result of existing works, we obtain a performance bound of BC with a possible solution to handle the quantity-quality dilemma.

\begin{theorem}[Performance bound of BC]\label{thm:1}
Let $\Pi$ be the set of all deterministic policy and $|\Pi|=|A|^{|S|}$. Assume that there does not exist a policy $\pi\in\Pi$ such that $\pi(s^i)=a^i, \forall i\in\{1,\cdots,|\caD|\}$. Let $\hatpi_b$ be the empirical behavior policy as well as the corresponding state marginal occupancy is $\hatrhob$. Suppose BC begins from initial policy $\pi_0$, and define $\rho_{\pi_0}$ similarly. Then, for any $\delta>0$, with probability at least $1-\delta$, the following inequality holds:
\begin{small}
\begin{equation}\label{eq:thm1}\begin{aligned}
    &\quad\quad\quad\quad\quad\quad\quad\quad\quad\quad\quad\quad\TV(\rho_{\pi}(s,a)\|\rho_{\pi_b}(s,a)) \le c(\pi_0, \pi_b, |\caD|)\\
    &\quad\quad\text{where} \quad c(\pi_0, \pi_b, |\caD|) = \frac{1}{2}\sum_{s\notin\caD}\rho_{\pi_b}(s) + \frac{1}{2}\sum_{s\notin\caD}|\rho_\pi(s)-\rho_{\pi_0}(s)| + \underbrace{ \frac{1}{2}\sum_{s\notin\caD}|\rho_{\pi_0}(s)-\rho_{\pi_b}(s)| }_{\text{initialization gap}} \\
    &+ \underbrace{ \frac{1}{2}\sum_{s\in\caD}|\rho_\pi(s)-\hatrhob(s)| + \frac{1}{|\caD|}\sum_{i=1}^{|\caD|} \sI\left[ \pi(s^i)\neq a^i \right]}_{\text{BC gap}} + \underbrace{ \left[\frac{\log|\caS|+\log(2/\delta)}{2|\caD|}\right]^{\frac{1}{2}} + \left[\frac{\log|\Pi|+\log(2/\delta)}{2|\caD|}\right]^{\frac{1}{2}} }_{\text{data gap}}
\end{aligned}\end{equation}
\end{small}
\end{theorem}

The proof can be found in \ap{ap:proof}. \theo{thm:1} shows the upper bound of the state-action distribution between the imitating policy $\pi$ and the behavior policy $\pi_b$, which consists of three important terms: the \textit{initialization gap}, the \textit{BC gap} and the \textit{data gap}. Specifically, the \textit{BC gap} arises from the empirical error and the difference between the imitating policy and the empirical behavior policy, which is corresponding to the training procedure. The \textit{data gap}, however, depends on the number of samples and complexity of the state space, acting as an intrinsic gap due to the dataset and the environment. 
As for the \textit{initialization gap}, it is in the form of distance between the state marginal distribution of the initial policy $\pi_0$ and behavior policy $\pi_b$ out of the dataset. Notice that the second term in \eq{eq:thm1} relates to the distance between the state marginal of the initial policy $\pi_0$ and the learned policy $\pi_b$ outside the data support, which is hard to measure theoretically due to the Markov property of the environment dynamics. Therefore, we estimate the empirical discrepancy outside the dataset $\frac{1}{2}\sum_{s\notin\caD}|\hat{\rho}_\pi(s)-\hat{\rho}_{\pi_0}(s)|$ for this term\footnote{For implementation details, see \ap{ap:implement-dis}.}. The results shown in \fig{fig:motivation-subb}, as expected, suggests that the second term in \eq{eq:thm1} in fact decreases as the initialized policy gets close to the demonstrated policy because of the poor generalization on unseen states, and the error can be further reduced with a larger dataset.
% due to the Markov property of the environment dynamics. However, we can not expect the agent to reach the unseen states and therefore we would like to assume little effect for this term. 
% Therefore, to imitate to the behavior policy, a farther initial policy requires more samples than a closer policy, as shown in \se{subsec:bc}.

Such analysis brings a possible theoretical explanation to our empirical intuition \se{subsec:bc}. Generally, given the same discrepancy $c(\pi_0, \pi_b, |\caD|)=C$, if the initialized policy narrows down the \textit{initialization gap} as is close to the demonstrated policy, then the requirement for more samples to minimize the \textit{data gap} can be relaxed. This may seem unreasonable in the learning theory in the traditional supervised learning domain. However, under the RL scenario, the performance of a policy depends on the accumulated reward along the rollout trajectories, which will lead to serious compounding error problems~\cite{ross2011reduction,ross2010efficient}. Therefore, as the distance between the initialized policy and the demonstrated policy gets closer, the generalization errors of the learned policy can be reduced.

% Correspondingly, a BC agent trained on a large amount of samples collected by a mixed of behavior policies can only learn a mediocre policy due to the minimization of the \textit{BC gap}. On the other hand, a random initialized agent initialized can not imitate high quality experience because of the \textit{data gap}. \minghuan{I think there is a problem: It is worth noting that if starting from a policy that is close to the behavior policy, the BC gap is small and the bound of data gap is rather loose, allowing BC to work well under this circumstance, as the motivated example shows.}

\textbf{Brief conclusion.} Both the experimental and the theoretical results indicate an interesting fact that the asymptotic performance of BC is highly related to the discrepancy between the initialized policy and the demonstrated policy. Specifically, a close-to-demonstration policy can easily imitate the demonstrated policy with fewer samples. On the contrary, when the distance between the initialized policy and the demonstrated policy is far, then successfully mimicking the policy will require much more samples. Such an observation motivates the intuition for proposing our Curriculum Offline Imitation Learning (COIL) in the following literature. The key insight enabling COIL is adaptively imitating the close policies with a small number of samples and finally terminates with the optimal behavior policy of the dataset.

\section{Curriculum Offline Imitation Learning}
\label{sec:method}
\subsection{Online RL as Imitating Optimal Policies}
\label{sec:online-offline}

\begin{figure}[!t]
\centering
\begin{subfigure}[b]{0.42\textwidth}
\includegraphics[width=0.9\linewidth]{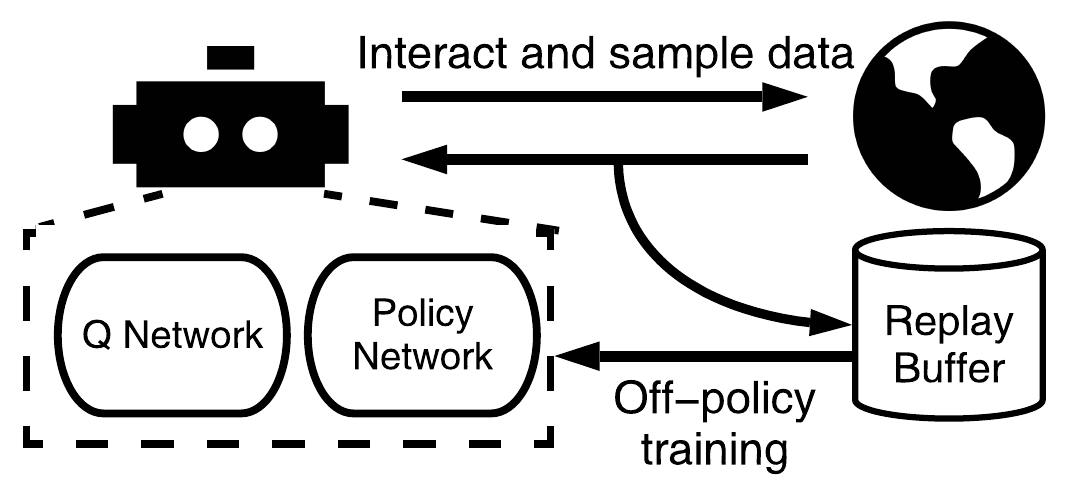}
\vspace{-8pt}
\label{fig:online-rl}
\caption{Online off-policy training.}
\end{subfigure}
\begin{subfigure}[b]{0.42\textwidth}
\label{fig:offline-rl}
\includegraphics[width=0.9\linewidth]{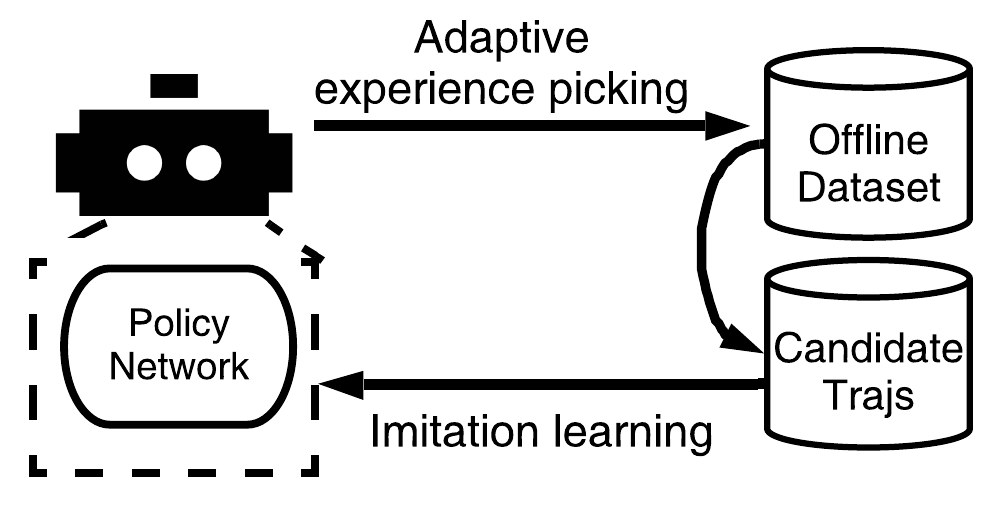}
\vspace{-8pt}
\caption{Curriculum offline imitation learning.}
\end{subfigure}
\vspace{-7pt}
\caption{Comparison between online off-policy training and curriculum offline imitation learning.}
\vspace{-10pt}
\label{fig:online-offline}
\end{figure}

Before starting to formulate our methodology, we first introduce the formulation of the online RL training. Typically, if we treat an optimization step of the policy as a training stage, an RL online learning algorithm can be realized as on-policy / off-policy depending on whether the agent is trained using the data collected by policies in the previous training stage. Taking off-policy RL as an example: beginning with a randomized policy $\pi^0$, at every training stage $i$, the agent uses its policy $\pi^i$ to interact with the environment to collect trajectory $\tau^i$ and save it to replay buffer $\caB$. The agent then samples several state-action pairs from $\caB$ and take an optimization step to get policy $\pi^{i+1}$, towards obtaining the most accumulated rewards:
\begin{equation}
    \mathop{\text{maximize}}_{\pi} \bbE_{\tau\sim\pi}[R(\tau)]~.
\end{equation}
Under the principle of maximum entropy, we can model the distribution of trajectories sampled by the optimal policy as a Boltzmann distribution~\cite{ziebart2008maximum,finn2016guided} as:
\begin{equation}\label{eq:max-ent-irl}
    P^*(\tau) \propto \exp{(R(\tau))}
\end{equation}
With such a model, trajectories with higher rewards are exponentially more preferred. And finding the optimal policy through RL is equivalent to imitating the optimal policy modeled by \eq{eq:max-ent-irl}~\cite{finn2016guided,finn2016connection}:
\begin{equation}
    \mathop{\text{minimize}}_{\pi} D_{KL}(P_{\pi}(\tau)\|P^*(\tau))~,
\end{equation}
where $P_{\pi}(\tau)=\rho(s_0)\sum_{t=0}^T \caT(s_{t+1}|s_t,a_t)\pi(a_t|s_t)$ is the distribution of generating a trajectory $\tau$ according to policy $\pi$. Thus, $\pi^{i+1}$ is updated follows the direction of minimizing the KL divergence:
\begin{equation}
    \pi^{i+1} = \pi^i - \nabla_{\pi} D_{KL}(P_{\pi}(\tau)\|P^*(\tau))
\end{equation}

\subsection{Offline RL as Adaptive Imitation}
\label{se:orl}

Compared with online policy training, the offline agent can only have a pre-collected dataset for policy training. Such dataset could be generated by a single policy, or collected by kinds of policies. In analogy to online RL, a similar solution on offline tasks can be imitating the optimal behavior policy from the dataset in an offline way. However, as we show before, offline IL methods, like BC, are only capable of matching the performance of the behavior policy, but hard to reach a good performance on the mixed dataset due to the quantity-quality dilemma.
% \paragraph{BC is acceptable for single-behavior dataset.} Direct BC from single-behavior dataset is able to always achieve the performance of the behavior policy, as Qin et al.~\cite{qin2021near} and our experiment suggest. Except discussing the expert policy, if the behavior data tends to be random, it is nearly impossible to learn a good policy from it. 
% In addition, for medium behavior dataset, imitating a medium behavior is also accessible, at least as pretraining for possible improvement.
% \paragraph{BC has quantity-quality dilemma on mixed dataset.} The most challenging and practical problem for BC will be a mixed dataset which contains the collected data from diverse behavior policies. In that case, simply BC from all data will lead to a mediocre policy that performs better than the worst policy and worse than the best policy. As we show in \se{sec:intro}, such a quantity-quality dilemma exists because the quantity of the high-quality demonstration can not be satisfied, and thus limit the performance of BC.

\subsubsection{Leverage Behavior Cloning with Curriculum Learning} 
In \se{sec:motivation} we have seen evidence that a possible solution to the quantity-quality dilemma could be adaptive imitation through the offline dataset. An overview of such an adaptive imitation learning diagram compared with online RL is shown in \fig{fig:online-offline}. Formally, with dataset $\caD=\{\tau\}_{1}^N$, at every training stage $i$, the agent updates its policy $\pi^i$ by adaptively selecting $\tau\sim\tilde{\pi}^i$ from $\caD$ as the imitating target such that:
\begin{equation}\label{eq:offline-update}
\begin{aligned}
    \pi^{i+1} =~ &\pi^i - \nabla_{\pi} D_{KL}(P_{\tilde{\pi}^i}(\tau)\|P_{\pi}(\tau))\\
    \text{s.t.~} %&D_{KL}(P_{\tilde{\pi}}(\tau)\|P_{\pi}(\tau)) \leq \epsilon\\
    &\bbE_{\tilde{\pi}}\left[D_{KL}(\tilde{\pi}^i(\cdot|s)\|\pi^i(\cdot|s))\right] \leq \epsilon\\
    &R_{\tilde{\pi}}^i - R_{\pi}^i \geq \delta
\end{aligned}
\end{equation}
where $\epsilon$ and $\delta$ are small positive numbers that limit the difference between the demonstrated policy $\tilde{\pi}^i$ and the learner $\pi$, and prevent $\pi$ from learning poorly behaved policies. Correspondingly, each training iteration $i$ creates a curriculum automatically such that a task $G_i$ is to imitating the closet demonstrated policy $\tilde{\pi_i}$ with $\caL_i^\pi=D_{KL}(P_{\tilde{\pi}_i}(\tau)\|P_{\pi}(\tau))$, and the target task $G_t$ is to imitate the optimal policy $\tilde{\pi}^*$. Specifically, we aims to construct a finite sequence ${\pi^0, \tilde{\pi}^1, \pi^1, \tilde{\pi}^2,\pi^2, \cdots, \tilde{\pi}^N, \pi^N}$ such that $ \pi^{i} \preceq \pi^{i+1}$, where $\tilde{\pi}^i$ is characterized by its trajectory, and is picked from $\caD$ based on the current policy $\pi^{i-1}$; $\pi^{i}$ is the imitation result taken $\tilde{\pi}^i$ as the target policy.

In the following sections, we explain how our algorithm is designed to achieve \eq{eq:offline-update} that leads the target policy $\tilde{\pi}$ to finally collapse into the optimal behavior policy $\tilde{\pi}^*$, while solving the quality-quantity dilemma of BC to avoid getting a mediocre result.

\subsubsection{Adaptive Experience Picking by Neighboring Policy Assessment}

We first provide a practical solution to evaluate whether $\bbE_{\tilde{\pi}}\left[D_{KL}(\tilde{\pi}(\cdot|s)\|\pi(\cdot|s))\right] \leq \epsilon$. In other words, we want to know whether a trajectory $\tau_i\in\caD$ is sampled by a neighboring policy. This can be regarded as finding a policy whose importance sampling ratio is near to 1, and thus a lot of density ratio estimation works can be referred to~\cite{nachum2019dualdice,zhang2020gendice}. However, such estimation requires extra costs on training neural networks, and the estimation is inaccurate with fewer data points. Therefore, in this paper, we design a simple yet efficient neighboring policy assessment principle instead that brings the algorithm into practice.

We assume that each trajectory is sampled by a single policy. Beyond such a slight and practical assumption, let $\pi$ be the current policy and trajectory $\tau_{\tilde{\pi}}=\{(s_0,a_0,s'_0,r_0),\cdots,(s_h,a_h,s'_h,r_h)\}$ is collected by an unknown deterministic behavior policy $\tilde{\pi}$ with exploration noise such that $\bbE_{(s,a)\in\tau_{\tilde{\pi}}} [\log{\tilde{\pi}(a_t|s_t)}]\geq \log{(1-\beta)}$, where $\beta$ denote the portion of exploration. In this way, we find a practical solution that relaxes the KL-divergence constraint through an observation:
% In this way, we have that:
% \begin{equation}
% \begin{small}
% \begin{aligned}\label{eq:find-tau}
%   \bbE_{\tilde{\pi}}\left[\kld(\tilde{\pi}(\cdot|s)\|\pi(\cdot|s))\right] \leq \epsilon \Leftrightarrow \bbE_{\tilde{\pi}} \left [\log{\frac{\tilde{\pi}(a|s)}{\pi(a|s)}}\right] \leq \epsilon \Rightarrow \log{\bbE_{(s,a)\in \tau_{\tilde{\pi}}} \left [\pi(a|s)\right]} \geq \log{(1-\beta)} - \epsilon
% \end{aligned}
% \end{small}
% \end{equation}
\begin{observation}\label{obs:find-tau}
Under the assumption that each trajectory $\tau_{\tilde{\pi}}$ in the dataset $\caD$ is collected by an unknown deterministic behavior policy $\tilde{\pi}$ with an exploration ratio $\beta$. The requirement of the KL divergence constraint $\bbE_{\tilde{\pi}}\left[\kld(\tilde{\pi}(\cdot|s)\|\pi(\cdot|s))\right] \leq \epsilon$ suffices to finding a trajectory that at least $1-\beta$ state-action pairs are sampled by the current policy $\pi$ with a probability of more than $\epsilon_c$ such that $\epsilon_c\geq 1 / \exp{\epsilon}$, \textit{i.e.}:
\begin{equation}\label{eq:find-tau-practice}
    \bbE_{(s,a)\in\tau_{\tilde\pi}}[\sI(\pi(a|s)\geq \epsilon_c)] \geq 1-\beta~,
\end{equation}
% where $\epsilon_c\geq 1 / \exp{\epsilon}$.
\end{observation}
The corresponding deviation is shown in \ap{proof:find-tau}. Therefore, to find whether $\tau_{\tilde{\pi}}$ is sampled by a neighboring policy, we calculate the probability of sampling $a_t$ at state $s_t$ by $\pi$ in $\tau$ for every timestep $\tau_{\tilde{\pi}}(\pi)=\{\pi(a_0|s_0), \cdots, \pi(a_h|s_h)\}$, where $h$ is the horizon of the trajectory. In practice, instead of fine-tuning $\epsilon$ and $\beta$, we heuristically set $\beta=0.05$ as an intuitive ratio of exploration. As for $\epsilon_c$, we let the agent choose the value through finding $N$ nearest policies that matches \eq{eq:find-tau-practice}.
% \begin{equation}
% \begin{aligned}
%   %\tau_{\tilde\pi} = \argmax_{\pi} (1-\beta) \frac{\sum_{t=0}^{h}\sI\left[\pi(a|s)\geq\epsilon\right]}{h}~.
% %   \tau_{\tilde\pi} = \argmax_{\tau=\{s_0,a_0,s_1,a_1,...,s_h,a_h\}}
%   \epsilon &= \argmax_{\epsilon}\frac{\sum_{t=0}^{h}\sI\left[\pi(a_t|s_t) \geq \epsilon\right]}{h}
%   &\text{s.t.} \quad\quad \frac{\sum_{t=0}^{h}\sI\left[\pi(a_t|s_t) \geq \epsilon\right]}{h} \geq 1-\beta~.
% \end{aligned}
% \end{equation}
% \hanye{Then we sort $\{\pi(a_i|s_i)\}_{i=0}^h$ in ascending order, denoting the result by $\{\pi(a'_i|s'_i)\}_{i=0}^h$, and regard the $\lfloor\beta h\rfloor$ smallest terms as exploration. }

\subsubsection{Return Filter}

We now present how to ensure the second constraint $R_{\tilde{\pi}} - R_{\pi} \geq \delta$, which is designed to refrain the performance from getting worse by imitating to a poorer target than the current level of the imitating policy. In a practical offline scenario, it is impossible to get the accurate return of the current policy, but we can evaluate its performance based on the current curriculum. To this end, we adopt a return filtering mechanism that filtrates the useless, poor-behaved trajectories.

In practice, we initialize the return filter $V$ with 0, and update the value at each curriculum. Specifically, if we choose $\{\tau\}_{1}^{n}$ from $\caD$ at iteration $k$, then $V$ is updated by moving average:
\begin{equation}
    V_k = (1-\alpha) \cdot V_{k-1} + \alpha \cdot \min\{R(\tau)\}_{1}^n
\end{equation}
where $\{R(\tau)\}_{1}^n$ is the accumulated reward set of trajectories $\{\tau\}_{1}^{n}$, and $\alpha$ is the moving window determining the filtering rate. Then, the dataset is updated as $\caD=\{\tau\in\caD\mid R(\tau)\ge V\}$.

\subsubsection{Overall Algorithm}
Combining the adaptive experience picking strategy and the return filter, we finally get the simple and practical curriculum offline imitation learning (COIL) algorithm. To be specific, COIL holds an experience pool that contains the candidate trajectories to be selected. Every training time creates a stage of the curriculum where the agent selects appropriate trajectories as the imitation target from the pool and learns them via direct BC. After training, the used experience will be cleaned from the pool, and the return filter also filtrates a set of trajectories. 
An attractive property of COIL is that it has a terminating condition that stops the algorithm automatically with a good policy when there is no candidate trajectory to be selected. This makes it easier to be applied in real-world applications without further finding the best-learned policy checkpoints under online evaluation as the previous algorithms do.
The step-by-step algorithm is shown in \alg{alg:coil}.

It is worth noting that COIL has only two critical hyperparameters, namely, the number of selected trajectories $N$ and the moving window of the return filter $\alpha$, both of which can be determined by the property of the dataset. Specifically, $N$ is related to the average discrepancy between the sampling policies in the dataset; $\alpha$ is influenced by the changes of the return of the trajectories contained in the dataset. % for better adaptation.
In the ablation study~\se{exp:ablation} and \ap{ap:ablation}, we demonstrate how we select different hyperparameters for different datasets.

\section{Related Work}

As a long-studied branch of RL, numerous solutions have been developed for offline RL with both model-free~\cite{fujimoto2019off,kumar2019stabilizing} and model-based~\cite{yu2020mopo,kidambi2020morel} algorithms. We briefly discuss the former from two categories, including RL-based methods and imitation-based methods.

\textbf{RL-based methods.} A naive thought is to apply off-policy RL algorithms such as SAC~\cite{haarnoja2018soft} and DDPG~\cite{lillicrap2015continuous} directly. However, as previous researchers~\cite{fujimoto2019off,kumar2019stabilizing} reveal, those online algorithms fail to work due to the severe extrapolation error or out-of-distribution problem. Thence, kinds of specifically designed algorithms have been proposed for offline RL. For instance, BCQ~\cite{fujimoto2019off} adopts a perturbation model to disturb the action sampled by a BC module to conduct Q-learning on the offline data. BEAR~\cite{kumar2019stabilizing} augments constraints on the policy to avoid out-of-distribution actions, over which it maximizes the approximate Q function.
The current state-of-the-art algorithm CQL~\cite{kumar2020conservative} performs strict constraints on the Q-function to learn an expectation lower bound of the true value, avoiding overestimation on out-of-distribution data. Typical drawbacks for these RL-based algorithms are serious bootstrapping errors and training instability that requires careful hyperparameter tuning with online evaluation. In comparison, imitation-based methods offer a candidate to mimic the demonstration.

\textbf{Imitation-based methods.} Another inspiration grows from the technique of offline imitation learning on how to learn from the demonstration. Most of these methods take the idea of behavior cloning (BC)~\cite{bain1995framework} that utilizes supervised learning to learn to act. However, due to the quantity-quality dilemma, BC limits mediocre performance on many datasets with mixed samples. Therefore,
BAIL~\cite{chen2020bail} constructs the upper-envelope on the value of the data and selects the best actions to imitate at each state and learn the policy based on BC. The main problem in BAIL is the requirement of regressing the value function with tricks to compute a value in the infinite horizon and the hyperparameter threshold on determining the best action.
ABM~\cite{siegel2020keep}, MARWIL~\cite{wang2018exponentially} and AWR~\cite{peng2019advantage} all take the idea of using an exponentially weighted version of BC, where the weights are determined by different forms of advantage function. These methods also require estimating the value function based on the offline data, which can be unstable and need many hyperparameters to control the learning procedure. Besides, it is also hard to tune an appropriate scale for the advantage for imitating the behavior policy. Compared with these methods, our COIL only has few important hyperparameters to be tuned without regressing any value function to achieve a stable performance.
\section{Experiments}
\label{sec:experiment}
Alongside the simple algorithm, we surprisingly find that COIL not only alleviates the quantity-quality dilemma but also achieves efficient and stable performance against competitive offline methods. Furthermore, we carry out a comprehensive analysis of the algorithm behind the phenomenon.

% We conduct several experiments to investigate the following research questions:
% \begin{enumerate}
%     \item[\textbf{RQ1}]  Is COIL able to  online training experience?
%     % \vspace{-4pt}
%     \item[\textbf{RQ2}] Does COIL works well on types of offline datasets?
%     % \vspace{-4pt}
%     \item[\textbf{RQ3}] What is the key ingredients of the proposed algorithm?
%     % \vspace{-8pt}
% \end{enumerate}

% To answer RQ1, we evaluate on training experience collected by an SAC agent during its training procedure.
% Regarding RQ2, we compare COIL with kinds of methods on a commonly used benchmark -- D4RL~\cite{fu2020d4rl} against many previous baselines.
% Finally, we conduct ablation studies on important hyperparameters for COIL.
% Due to the space limit, we leave experiment details, additional results and ablation study in Appendix.

\subsection{Offline Learning from Online Learning Experience}
\label{se:exp1}

\begin{figure}[tb]
\vspace{-15pt}

\begin{subfigure}[b]{0.3\textwidth}
\centering
\includegraphics[height=0.86\linewidth]{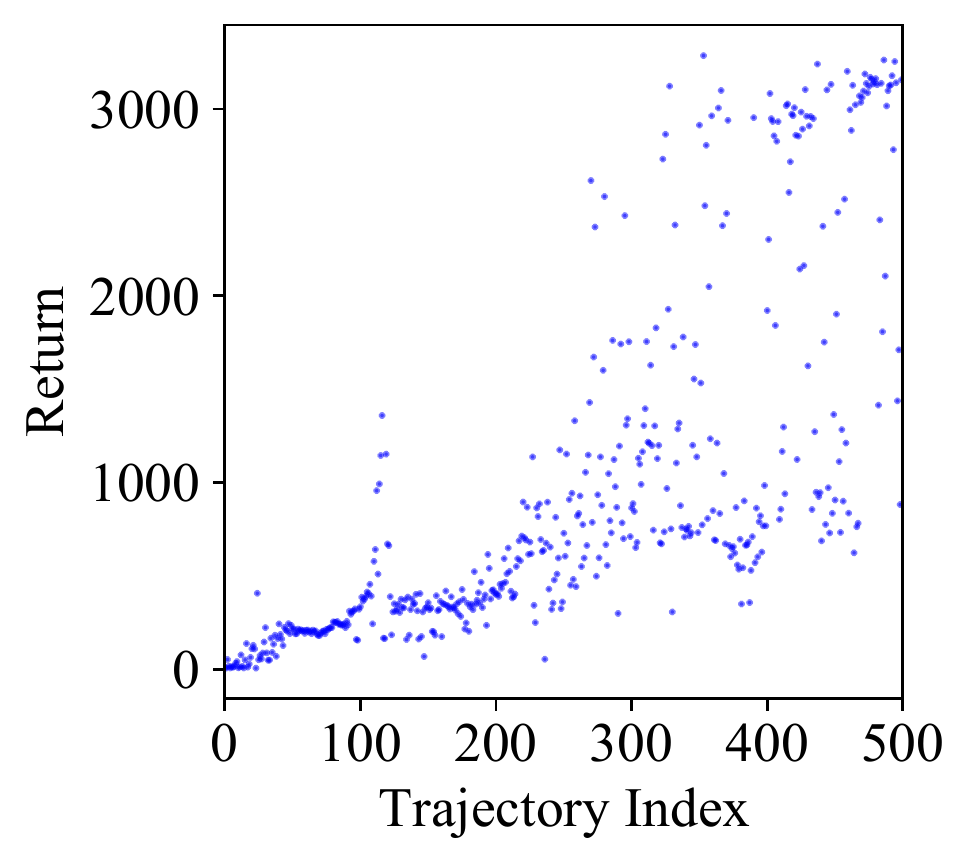}

\vspace{-5pt}
\caption{Hopper-final.}
\label{fig:hopper-final}
\end{subfigure}
\begin{subfigure}[b]{0.3\textwidth}
\centering
\includegraphics[height=0.86\linewidth]{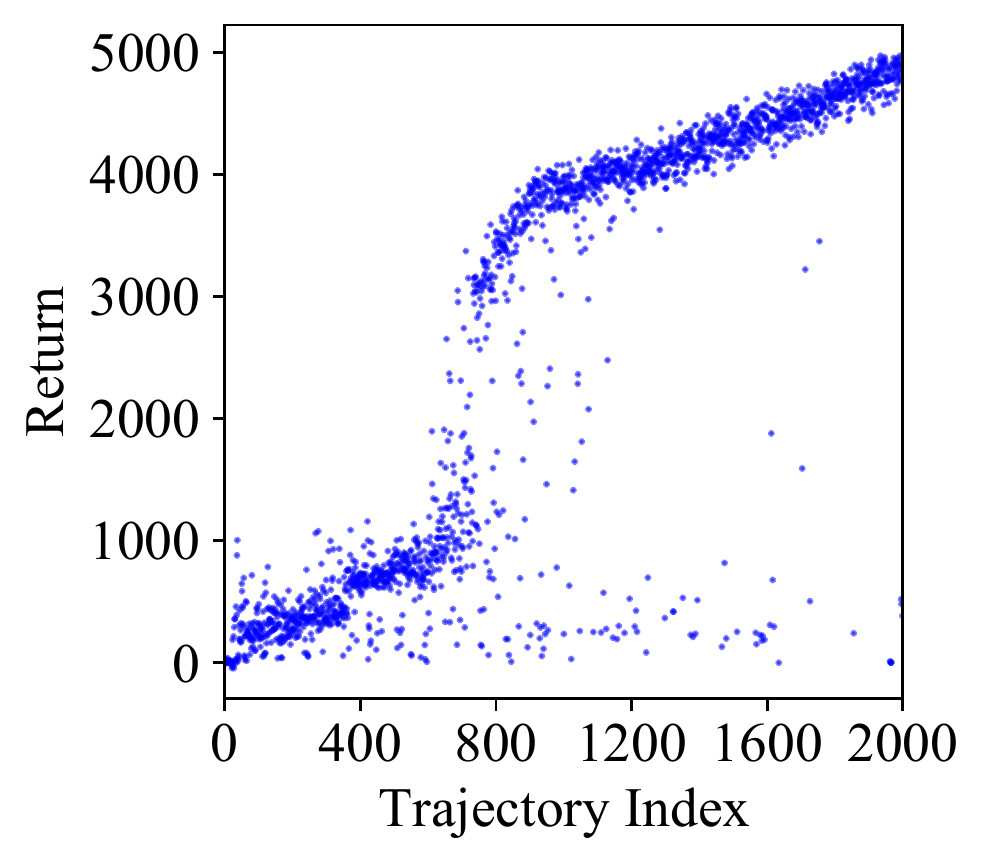}

\vspace{-5pt} 
\caption{Walker2d-final.}
\label{fig:walker2d-final}
\end{subfigure}
\begin{subfigure}[b]{0.3\textwidth}
\centering
\includegraphics[height=0.86\linewidth]{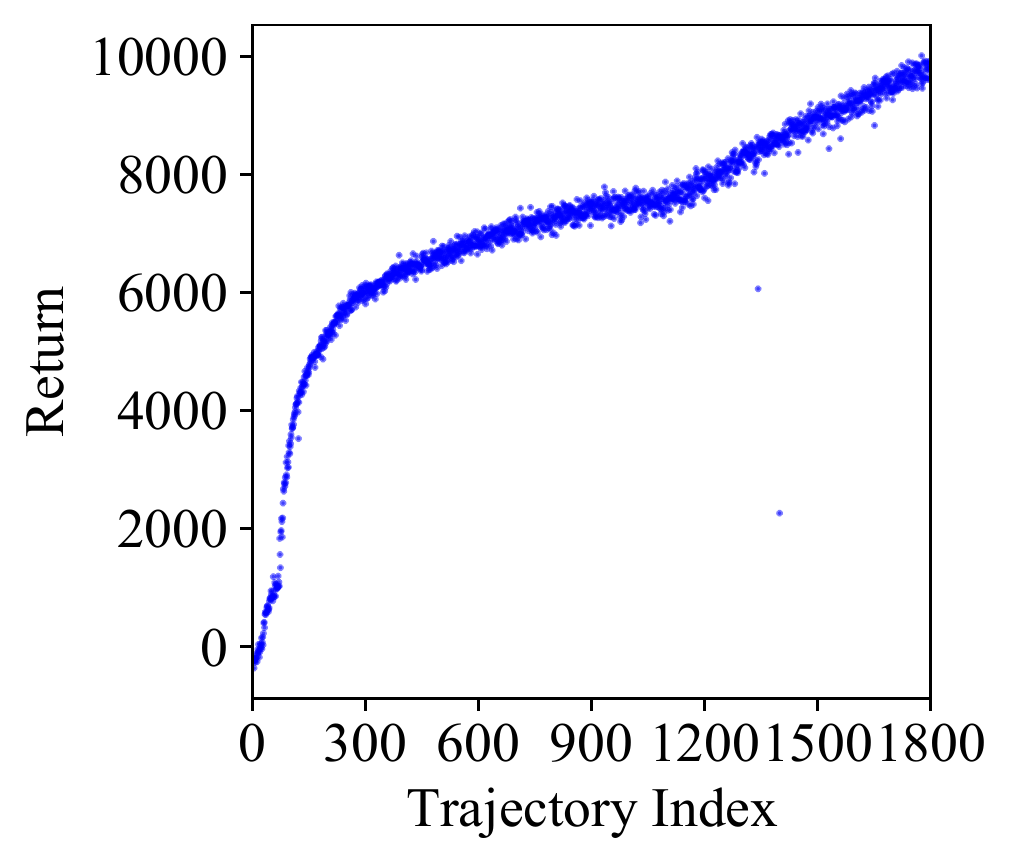}

\vspace{-5pt} 
\caption{HalfCheetah-final.}
\label{fig:halfcheetah-final}
\end{subfigure}

\vspace{-5pt}
\caption{Trajectories in final datasets, sorted in the online training order.}

\label{fig:finaldataset}
\vspace{-15pt}
\end{figure}

% It is worth noting that similar datasets have been already studied by previous works, such as Fujimoto et. al~\cite{fujimoto2019off}, where they call it the \textit{final buffer} and show that both off-policy RL and BC fail.
We are curious about the performance of COIL on the \textit{final buffer} datasets containing a complete training experience of an online agent~\cite{fujimoto2019off}, because such dataset is mixed with various levels of policies. Therefore direct BC will easily fail as shown in \se{sec:intro}. 
In our case, we first train an SAC agent from scratch to convergence and save all the training experience from its interactions with the environment, and therefore including exploration actions. Specifically, we conduct our experiments on three continuous control tasks: Hopper, Walker2d, and HalfCheetah. \fig{fig:finaldataset} illustrates trajectories in each dataset, sorted in the collected order.

To show the effectiveness of COIL, we compare it with several strong baselines, including the state-of-the-art offline RL algorithm CQL~\cite{kumar2020conservative}, and imitation-based methods AWR~\cite{peng2019advantage} and BAIL~\cite{chen2020bail}. For CQL, AWR and BAIL we use their open-source implementation with their default hyperparameter, and for BC we test the same implementation as COIL.
The quantitative converged results are listed in \tb{tab:final}, from which we observe that COIL substantially outperforms the other baselines for the \textit{final buffer} dataset. Also, BAIL and AWR can not always find the optimal behavior due to the difficulty of its hyperparameters tuning and value regression.
Specifically, compared with BC that learns a mediocre policy, COIL reaches the performance close to the optimal policy.

\begin{figure}[htbp]
\centering
\begin{minipage}{0.95\linewidth}
\vspace{-9pt}
\begin{subfigure}[b]{0.3\textwidth}
\centering
\includegraphics[width=0.87\linewidth,height=0.85\linewidth]{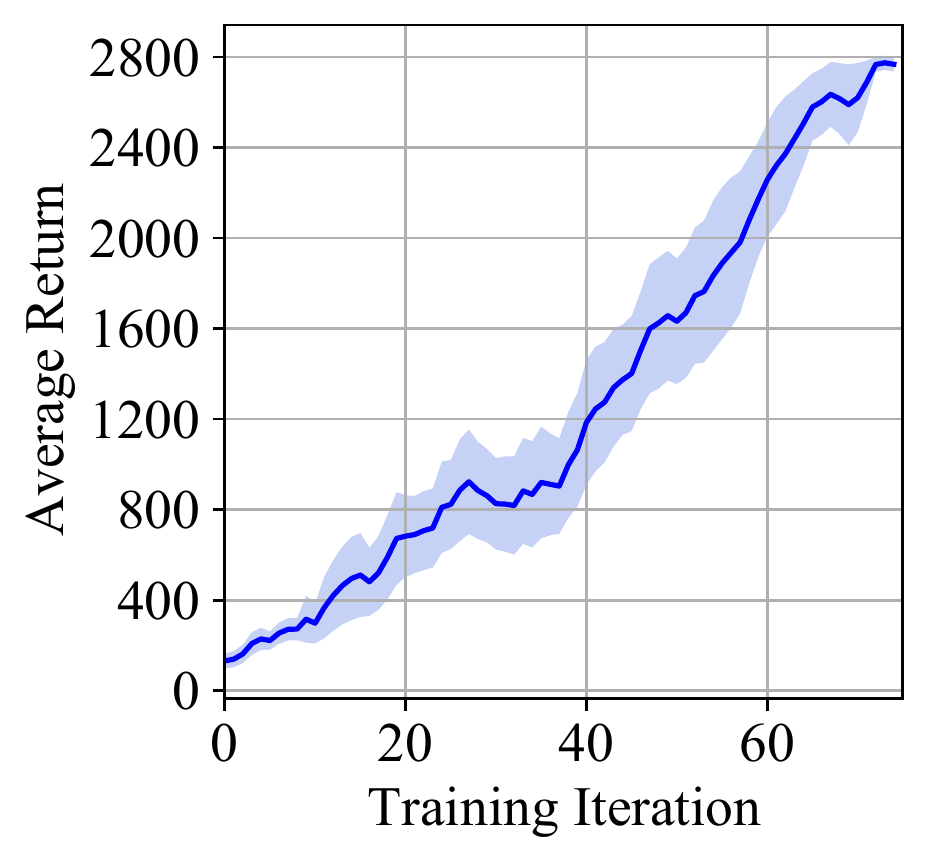}

\vspace{-5pt}
%\caption{Hopper-final.}
\label{fig:hopper-final-return}
\end{subfigure}
\begin{subfigure}[b]{0.3\textwidth}
\centering
\includegraphics[width=0.87\linewidth,height=0.85\linewidth]{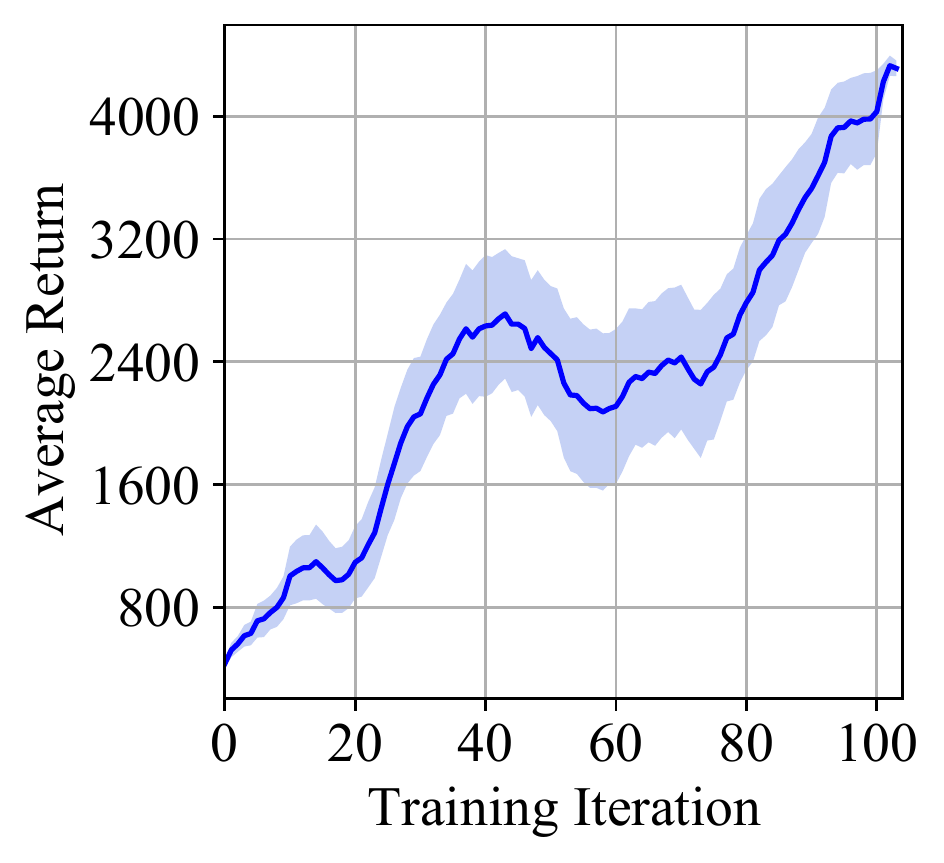}

\vspace{-5pt} 
%\caption{Walker2d-final.}
\label{fig:walker2d-final-return}
\end{subfigure}
\begin{subfigure}[b]{0.3\textwidth}
\centering
\includegraphics[width=0.87\linewidth,height=0.85\linewidth]{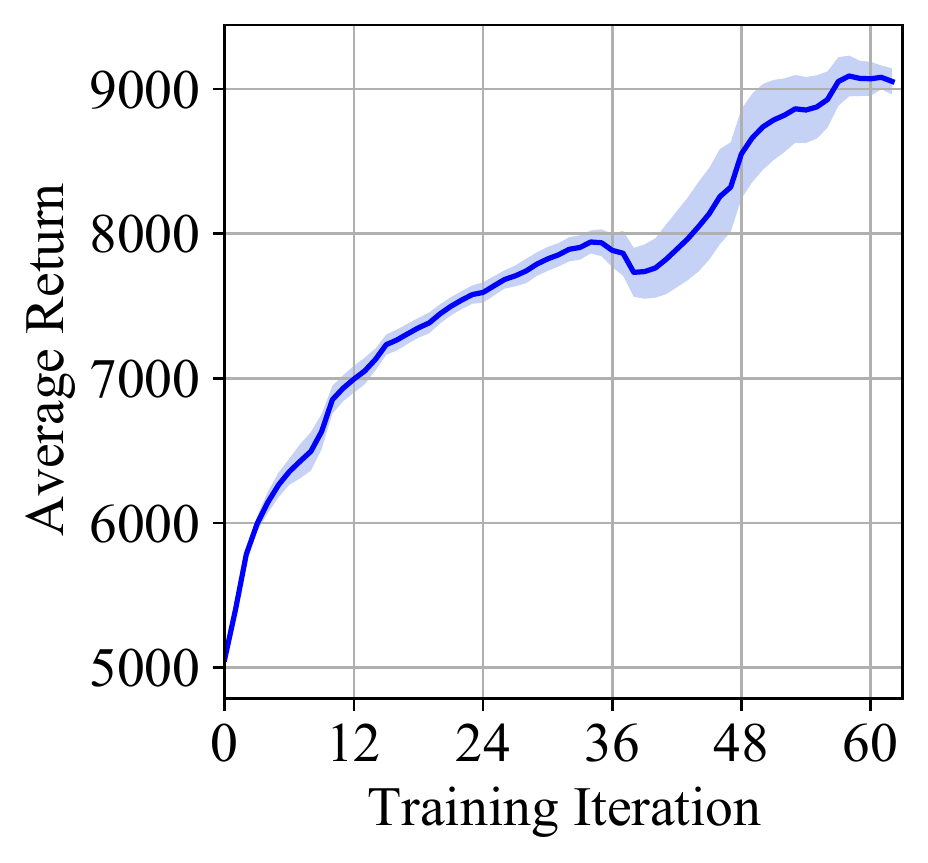}

\vspace{-5pt} 
%\caption{HalfCheetah-final.}
\label{fig:halfcheetah-final-return}
\end{subfigure}

\vspace{2pt}
\begin{subfigure}[b]{0.3\textwidth}
\centering
\includegraphics[width=0.87\linewidth,height=0.85\linewidth]{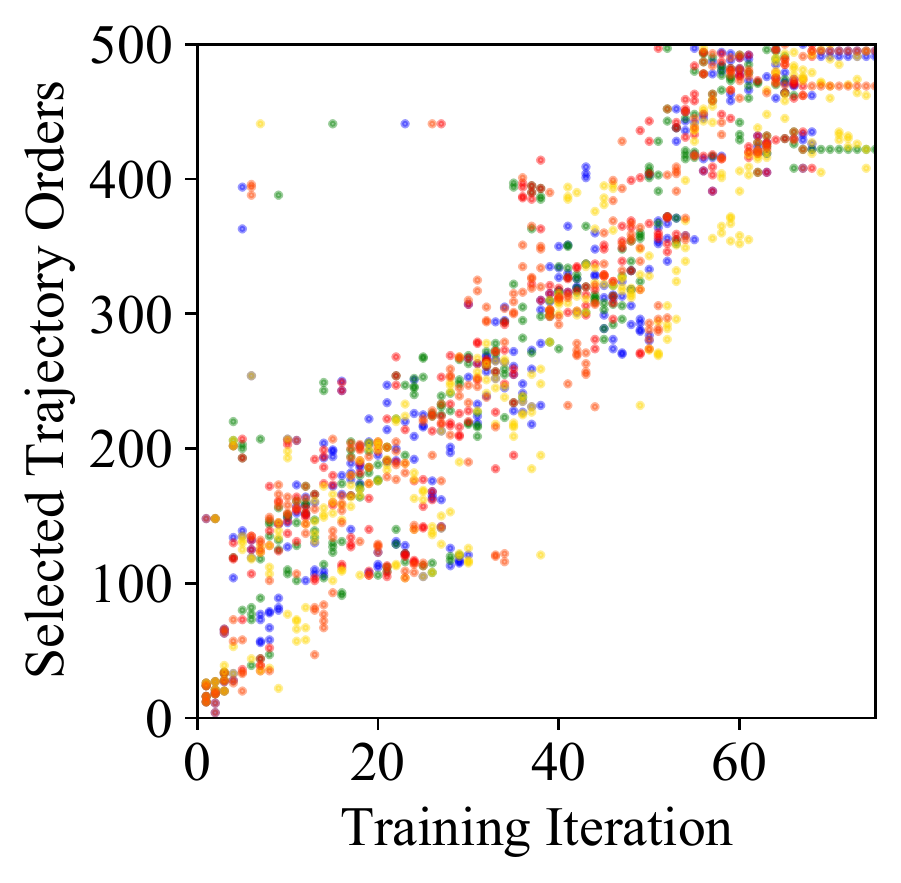}

\vspace{-6pt}
\caption{Hopper-final.}
\label{fig:hopper-final-traj}
\end{subfigure}
\begin{subfigure}[b]{0.3\textwidth}
\centering
\includegraphics[width=0.87\linewidth,height=0.85\linewidth]{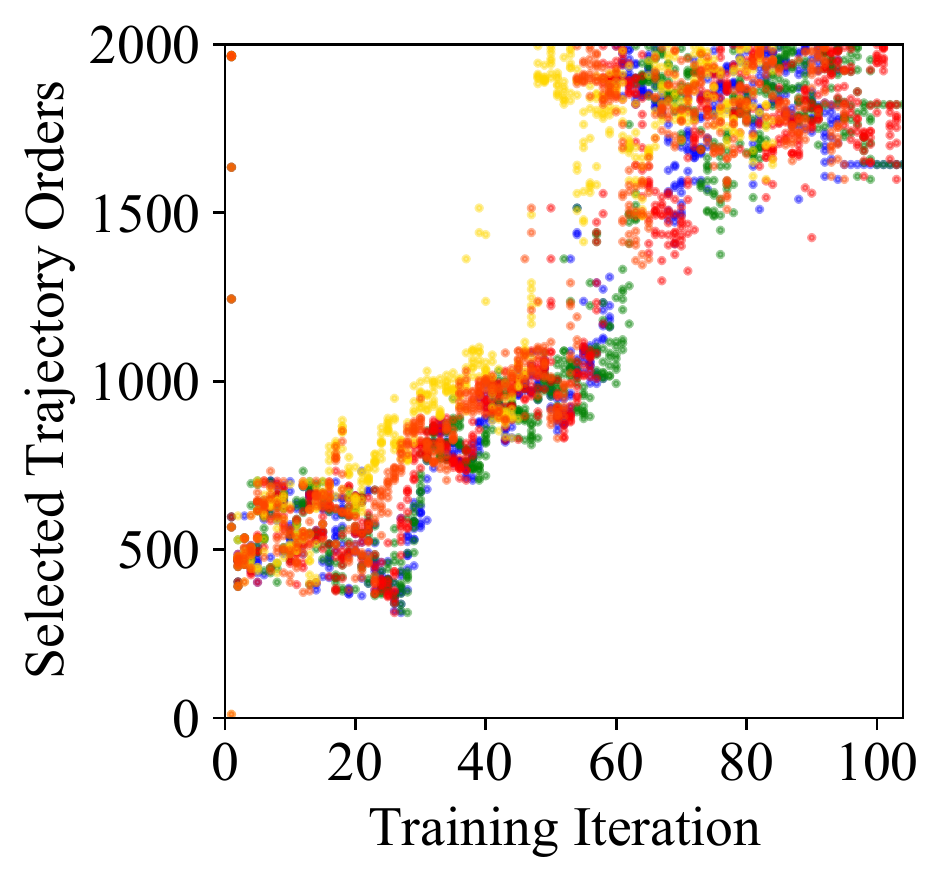}

\vspace{-6pt} 
\caption{Walker2d-final.}
\label{fig:walker2d-final-traj}
\end{subfigure}
\begin{subfigure}[b]{0.3\textwidth}
\centering
\includegraphics[width=0.87\linewidth,height=0.85\linewidth]{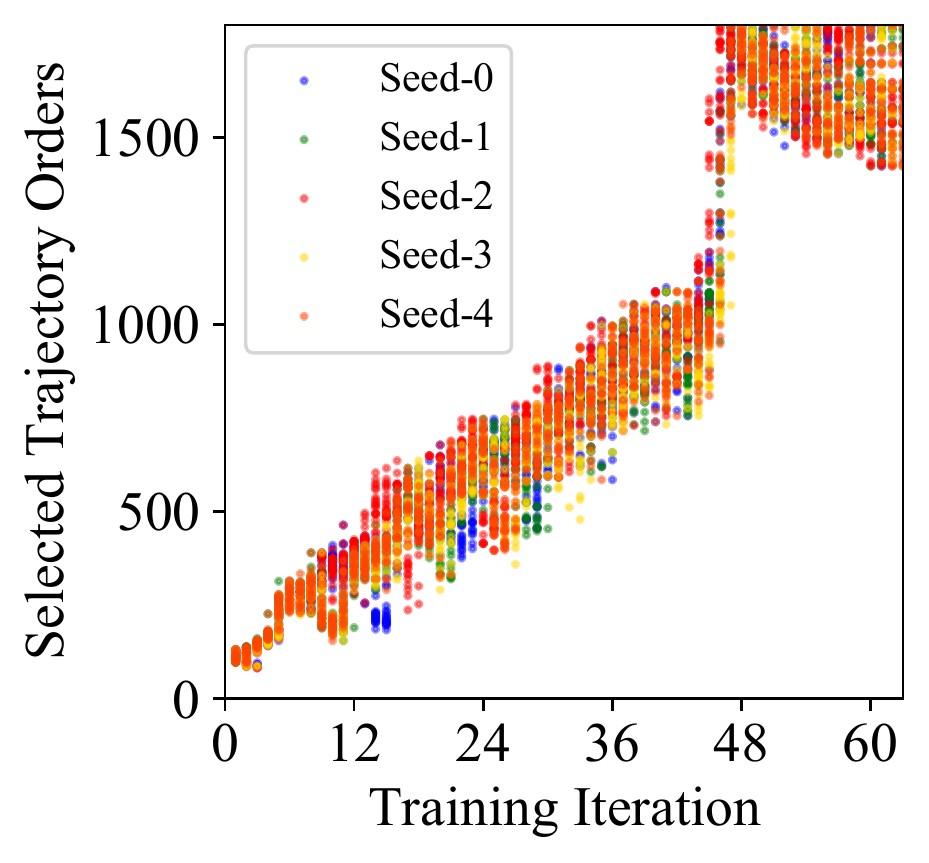}

\vspace{-6pt} 
\caption{HalfCheetah-final.}
\label{fig:halfcheetah-final-traj}
\end{subfigure}
\end{minipage}

\vspace{-7pt}
\caption{Training curves and orders of selected trajectories.}
\vspace{-10pt}
\label{fig:final-results}
\end{figure}

To further illustrate how our algorithm works, we also record the oracle online order of the trajectory sampled by the offline agent, as shown in \fig{fig:final-results}. Notably, COIL keeps a similar training path as the online agent, thanks to the experience picking strategy and the return filter. The corresponding learning curves shown in \fig{fig:final-results} are stable and following similar shapes as the ordered datasets. Notably, COIL finally terminates with a near-data-optimal policy, suggesting a nice property that the last offline model can be a great model for deployment, unlike previous algorithms that relies on online evaluation to select the best checkpoint.

\begin{table}[hbtp]
\centering
\vspace{-11pt}
\caption{Average performances on \textit{final} datasets, the means and standard deviations are calculated over 5 random seeds. \textit{Behavior} shows the average performance of the behavior policy that collects the data.}
\resizebox{1\columnwidth}{!}{
\begin{tabular}{cccccccc}
\toprule
\textbf{Dataset}           & \textbf{Expert (SAC)} & \textbf{Behavior} & \textbf{BC}       & \textbf{AWR}      & \textbf{BAIL}             & \textbf{CQL}                & \textbf{COIL (Ours)} \\
\midrule
\texttt{hopper-final}      & 3163.3 (44.4)       & 974.5             & 1480.4 (800.2)  & 1609.7 (489.7)  & 2296.9 (915.9)          & 501.5 (227.5)              & \textbf{2872.5 (133.9)} \\
\texttt{walker2d-final}    & 4866.03 (68.6)      & 2684.9            & 2099.6 (2101.3) & 3213.8 (1682.9) & \textbf{4236.2 (1531.1)}         & 2604.3 (1937.6)           & \textbf{4391.3 (697.8)} \\
\texttt{halfcheetah-final} & 9739.1 (113.6)      & 7122.4            & 6125.6 (3910.9) & 7600.9 (1153.4) & 9745.0 (880.3) & \textbf{10882.0 (1042.7)} & \textbf{9328.5 (1940.6)} \\
\bottomrule
\end{tabular}
}
\vspace{-9pt}
\label{tab:final}
\end{table}

\subsection{D4RL Benchmark}
\label{se:d4rl}

To further show the power of COIL, we conduct comparison experiments on a common-used D4RL benchmark~\cite{fu2020d4rl} in \tb{tab:d4rl}. For BC we include the BC results reported in Fu et al.~\cite{fu2020d4rl} (denoted as \textit{BC (D4RL)}) and our implementation of BC (denoted as \textit{BC (Ours)}). Besides, we compare BAIL~\cite{chen2020bail}, MOPO~\cite{yu2020mopo}, and the state-of-the-art results reported in Fu et al.~\cite{fu2020d4rl} (denoted as \textit{SoTA}).
% Specifically, we test COIL on four kinds of datasets: \texttt{random} is collected by a randomly initialized SAC agent without training, and \texttt{medium} by the SAC agent early-stopped on the half way of training. Under a different setting as mixed behaviour policies, \texttt{medium-replay} consists of all samples saved into the replay buffer during training an online agent until the agent achieves a mediocre performance, and \texttt{medium-expert} contains an equal mixture of trajectories samples by a medium level policy and an expert policy.
As our expected, BC is able to approach or outperform the performance of the behavior policy on the datasets generated from a single policy, marked as “\texttt{random}”, and “\texttt{medium}”, but still remains a gap between the optimal behavior policy (\textit{Best $1\%$} column). As a comparison, COIL achieves the performance of the optimal behavior policy on most datasets. As is noticed that doing so will allow COIL to beat or compete with the state-of-the-art results. To our surprise, on the \texttt{halfcheetah} domain, previous RL-based offline algorithms have the potential to surpass the optimal behavior policy (although with careful hyperparameter tuning), showing the advantage of RL-based learning. It is worth noting that for model-based algorithm like MOPO, it behaves well on the \texttt{medium-replay} datasets due to the sufficient data to learn a good environment model; but it can hardly outperform SoTA model-free results on other datasets.

\begin{table}[hbtp]
\centering
\vspace{-13pt}
\caption{Average performance on D4RL datasets. Results in \textcolor{lightgray}{gray} columns is our implementation that are tested among 5 random seeds. The other results are based on numbers reported in D4RL among 3 random seeds without standard deviations. \textit{Best 1\%} shows the average return of the top 1\% best trajectories, representing the performance of the optimal behavior policy; \textit{Behavior} shows the average performance of the dataset.}
\resizebox{1.0\columnwidth}{!}{
\begin{tabular}{lcccc>{\columncolor{lightgray}}c>{\columncolor{lightgray}}cccc}
\toprule
\textbf{Dataset}                   & \textbf{Expert (D4RL)} & \textbf{Behavior} & \textbf{Best 1\%} & \textbf{BC (D4RL)} & \textbf{BC (Ours)} & \textbf{COIL (Ours)}        & \textbf{BAIL} & \textbf{MOPO} & \textbf{SoTA (D4RL)}\\                                  
\midrule
\texttt{hopper-random}             & 3234.3                 & 295.1             & 340.4             & 299.4              & 330.1 (3.5)      & 378.5 (15.2)                  & 318.0 (5.1)   & \textbf{432.6}   & 376.3  \\
\texttt{hopper-medium}             & 3234.3                 & 1018.1            & 3076.4            & 923.5              & 1690.1 (852.0)   & \textbf{3012.0 (332.2)}       & 1571.5 (900.7) & 862.1          & 2557.3 \\
\texttt{hopper-medium-replay}      & 3234.3                 & 466.9             & 1224.8            & 364.4              & 853.6 (397.5)    & 1333.7 (271.1)                & 808.7 (192.5)  & \textbf{3009.6}  & 1227.3 \\
\texttt{hopper-medium-expert}      & 3234.3                 & 1846.8            & 3735.7            & \textbf{3621.2}    & \textbf{3527.4 (504.1)}   & \textbf{3615.5 (168.9)}                & 2435.9 (1265.2) & 1682.0  & 3588.5 \\
\midrule
\texttt{walker2d-random}           & 4592.3                 & 1.1               & 25.0              & 73.0               & 171.0 (59.3)     & 320.5 (70.7)     & 130.8 (87.2) & \textbf{597.1}  & 336.3 \\
\texttt{walker2d-medium}           & 4592.3                 & 496.4             & 3616.8            & 304.8              & 1521.9 (1381.3)  & 2184.5 (1279.2)           & 1242.4 (1545.7) & 643.0          & \textbf{3725.8} \\
\texttt{walker2d-medium-replay}    & 4592.3                 & 356.6             & 1593.7            & 518.6              & 715.0 (406.5)    & 1439.9 (347.0)   & 532.9 (359.0) & \textbf{1961.1}& 1227.3 \\
\texttt{walker2d-medium-expert}    & 4592.3                 & 1059.7            & 5133.4            & 297.0              & \textbf{3488.6 (1815.1)} & \textbf{4012.3 (1463.0)}  & \textbf{3633.9 (1839.7}) & 2526.0       & \textbf{5097.3}    \\
\midrule
\texttt{halfcheetah-random}        & 12135.0                & -302.6            & -85.4             & -17.9              & -124.3 (60.6)    & -0.3 (0.7)                &  -96.4 (49.7) & 3957.2        & \textbf{4114.8} \\
\texttt{halfcheetah-medium}        & 12135.0                & 3944.9            & 4327.7            & 4196.4             & 3276.4 (1500.7)  & 4319.6 (243.7)            & 4277.6 (564.9) & 4987.5        & \textbf{5473.8}  \\
\texttt{halfcheetah-medium-replay} & 12135.0                & 2298.2            & 4828.4            & 4492.1             & 4035.7 (365.4)   & 4812.0 (148.7)            & 3854.8 (966.3) & \textbf{6700.6} & 5640.6 \\
\texttt{halfcheetah-medium-expert} & 12135.0                & 8054.4            & 12765.4           & 4169.4             & 633.2 (2152.9)   & \textbf{10535.6 (3334.9)} & \textbf{9470.3 (4178.9)} & 7184.7        & 7750.8 \\ 
\bottomrule
\end{tabular}
}
\label{tab:d4rl}
\vspace{-12pt}
\end{table}

\subsection{Ablation Study}
\label{exp:ablation}

\begin{wrapfigure}{r}{0.5\textwidth}
\vspace{-30pt}
\centering
\includegraphics[width=0.97\linewidth]{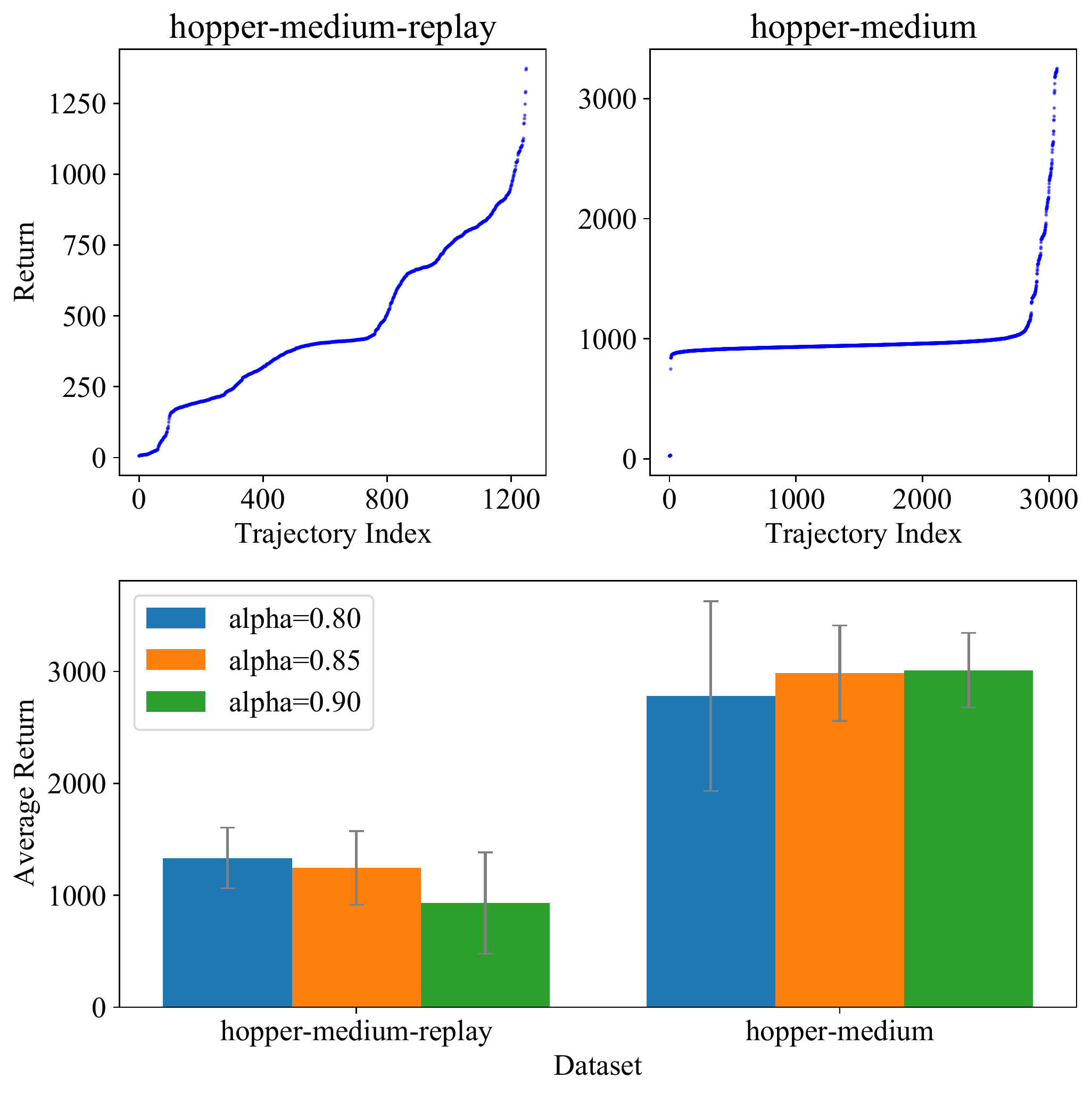}
\vspace{-6pt}
\caption{Returns of trajectories in \hytt{hopper-medium-replay} and \hytt{hopper-medium}, and final performances of COIL with different $\alpha$.}
\label{fig:ablation}
\end{wrapfigure}
The ablation study aims to illustrate how hyperparameters can be determined according to the property of the offline datasets without online evaluation. In particular, COIL has two critical hyperparameters: $\alpha$ and $N$. Since $\alpha$ determines the rate of filtering out the bad trajectories as the agent imitates a better policy where small $\alpha$ corresponds to a high filtering rate, it is highly influenced by the changes in returns of the trajectories in the dataset. Taking \hytt{hopper-medium-replay} and \hytt{hopper-medium} for examples, the former consists of online training experience, and the return of trajectories changes rapidly; on the contrary, most of the trajectories in the latter are at the same return level. Therefore, intuitively, a small value of $\alpha$ should be assigned to \hytt{hopper-medium-replay} while a large value to \hytt{hopper-medium}, as confirmed by the experimental results shown in \fig{fig:ablation}. 
Ablation experiments on the other hyperparameter $N$ are illustrated in \ap{ap:addition-exps}, saying that it can also be tuned according to the characterization of the dataset.

\subsection{Compared with Naive Curriculum Strategies}
\label{exp:naive-curriculum}

In this section, we propose comparing the curriculum strategy of COIL with naive curriculum strategies to determine whether COIL is a better choice. The first strategy is called \textit{return-ordered BC}, which conducts curriculum imitation learning based only on returns of the trajectories. Similar to COIL, it picks $N_{RBC}$ trajectories with the lowest returns for each curriculum to perform behavioral cloning, and then removes them from the dataset. When the dataset is empty, the algorithm stops. The other strategy is called \textit{buffer-shrinking BC}. As its name suggests, it conducts curriculum imitation learning by shrinking the dataset ordered by the return after training at each curriculum. In detail, \textit{buffer-shrinking BC} begin its training with the entire dataset in the buffer; after a fixed number of gradient steps, it shrinks the buffer by discarding $p\%$ of trajectories with the lowest returns; then the training is continued on the remaining trajectories. In our experiment, we choose $p=20$ so that the algorithm will stop after 5 times of shrinkage.

\begin{figure}[htbp]
\begin{subfigure}[b]{0.3\textwidth}
\centering
\includegraphics[height=0.86\linewidth]{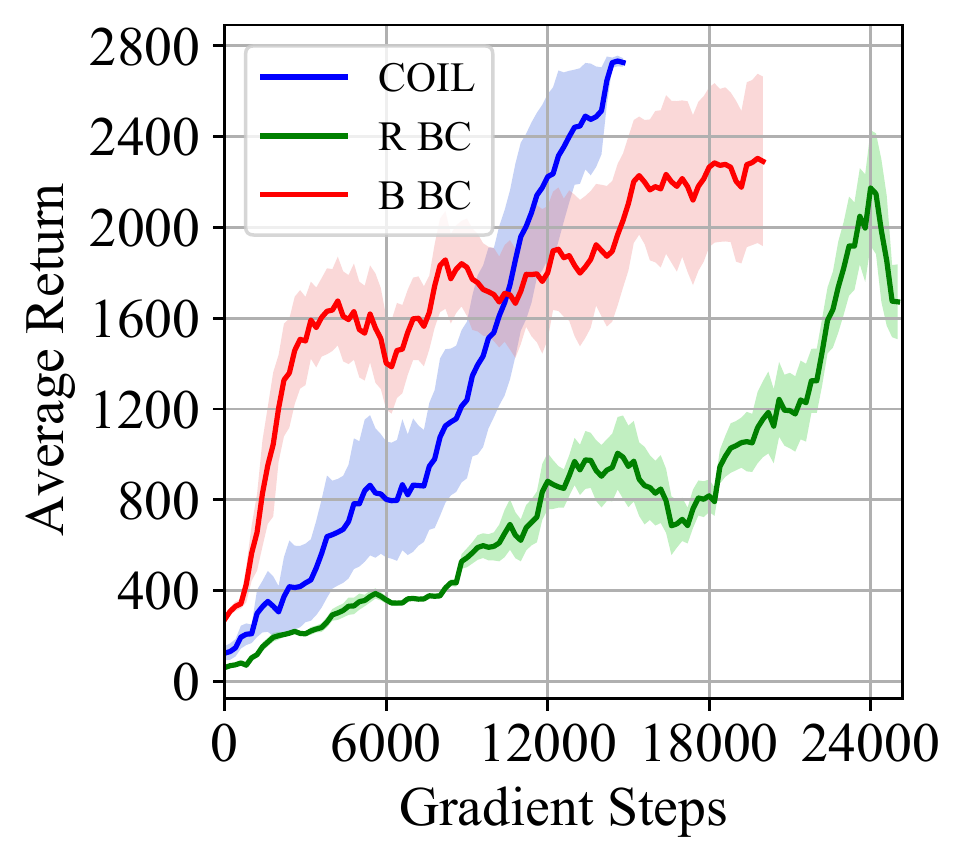}
\caption{Hopper-final.}
\label{fig:hopper-naive}
\end{subfigure}
\begin{subfigure}[b]{0.3\textwidth}
\centering
\includegraphics[height=0.86\linewidth]{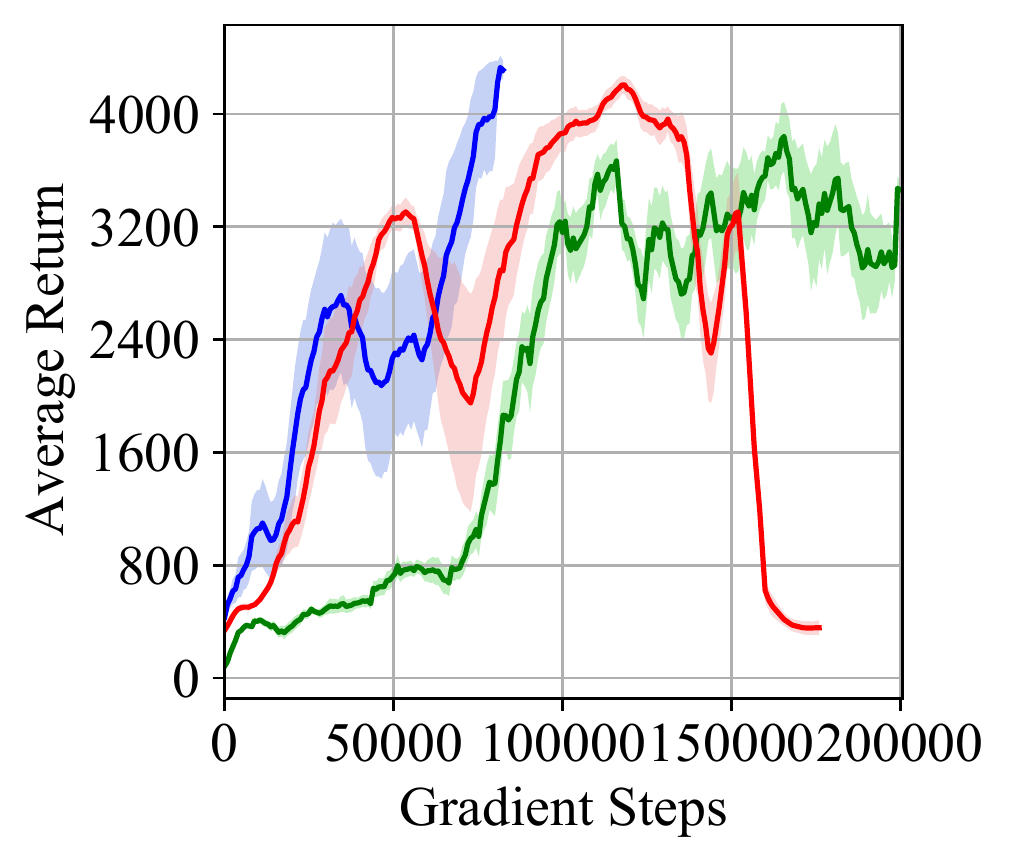}
\caption{Walker2d-final.}
\label{fig:walker2d-naive}
\end{subfigure}
\begin{subfigure}[b]{0.3\textwidth}
\centering
\includegraphics[height=0.86\linewidth]{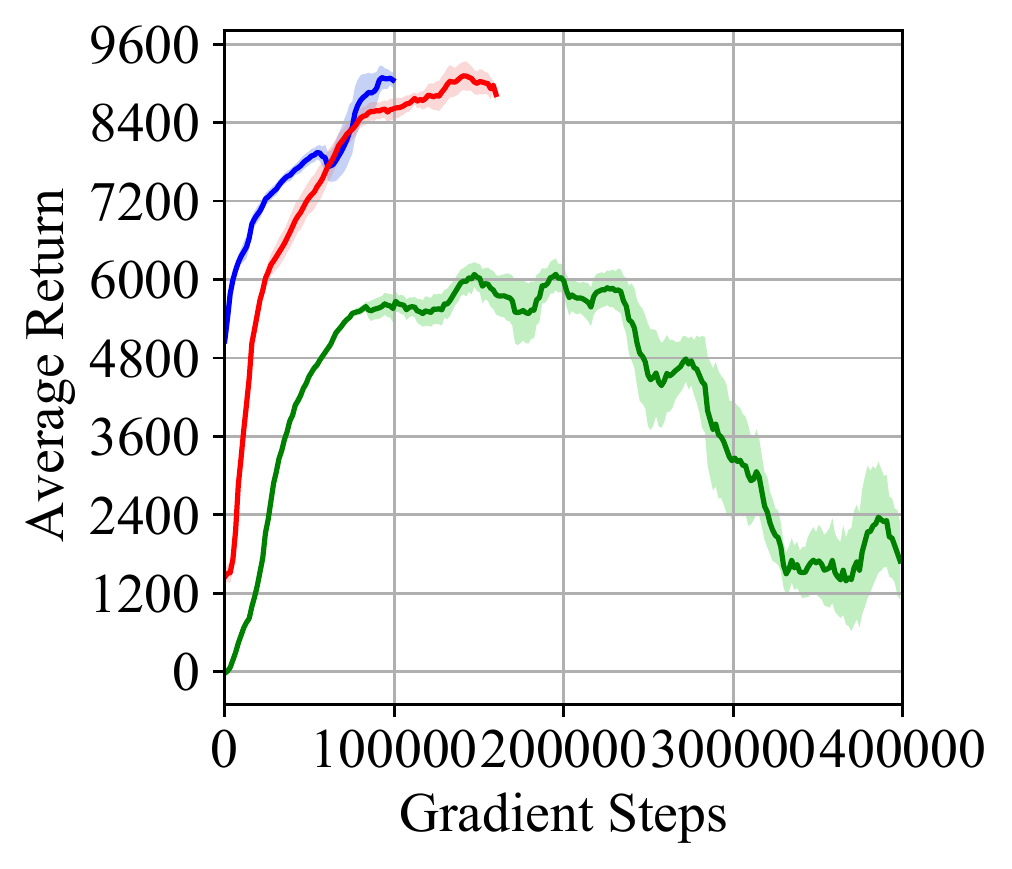}
\caption{HalfCheetah-final.}
\label{fig:halfcheetah-naive}
\end{subfigure}

\caption{Comparison of training curves between COIL and Return-ordered BC (\textit{R BC}) and Buffer-shrinking BC (\textit{B BC}) on \textit{final} datasets with the same batch size. Different strategies terminate with different gradient steps.}
\label{fig:final-naive}
\end{figure}

The learning results on \textit{final} datasets are shown in \fig{fig:final-naive}. As expected, \textit{return-ordered BC} leads to mediocre behaviors since similar returns do not always account for similar policies due to the exploration noise. In addition, \textit{buffer-shrinking BC} is not usually stable to achieve the optimal behavior. Since the shrink strategy is totally hand-crafted, the \textit{quantity-quality dilemma} is not eliminated. The mediocre trajectories in the last curriculum will lead to the failure. On the contrary, COIL succeeds in the optimal behavior policy with the highest training efficiency (the least gradient steps), indicating the advantage of the policy-distance-based curriculum.

\section{Conclusion}
\label{sec:conclusion}

In this paper, we analyze the quantity-quality dilemma of behavior cloning (BC) from both an experimental and a theoretical point of view, which motivates us to propose the curriculum offline imitation learning (COIL). COIL takes advantage of imitation learning by improving the current policy with adaptive neighboring policies. Experiments show good properties of COIL with competitive evaluation results against state-of-the-art offline RL algorithms. COIL may provide a practical way for bringing offline RL into practice due to its simplicity and effectiveness, but it also limits into the performance of the dataset.

% \clearpage
\section*{Acknowledgements}
The corresponding author Weinan Zhang is supported by ``New Generation of AI 2030'' Major Project (2018AAA0100900), Shanghai Municipal Science and Technology Major Project (2021SHZDZX0102) and National Natural Science Foundation of China (62076161, 61632017).
The author Minghuan Liu is supported by Wu Wen Jun Honorary Doctoral Scholarship, AI Institute, Shanghai Jiao Tong University. We sincerely thank the reviewers for helpful feedback. 
\bibliographystyle{plain}
\bibliography{ref}

\newpage
\appendix
\appendixpage

\section{Algorithm}\label{ap:algo}
\begin{algorithm}
    \caption{Curriculum Offline Imitation Learning (COIL)}
    \label{alg:coil}
    \begin{algorithmic}
        \REQUIRE Offline dataset $\caD$, number of trajectories picked at each curriculum $N$, moving window of the return filter $\alpha$, number of training iteration $L$, batch size $B$, number of pre-train times $T$, and the learning rate $\eta$.
        \STATE Initialize policy $\pi$ with random parameter $\theta$.
        \STATE Initialize the return filter $V=0$.
        \IF{$\caD$ is collected by a single policy}
            \STATE Do pre-training for $T$ times using BC.
        \ENDIF
        \WHILE{$\caD \neq \emptyset$}
            \FORALL{$\tau_i\in\caD$}
                \STATE Calculate $\tau_i(\pi)=\{\pi(a_0^i|s_0^i), \pi(a_1^i|s_1^i), \cdots, \pi(a_h^i|s_h^i)\}$.
                \STATE Sort $\tau_i(\pi)$ into $\{\pi(\bar{a}_0^i|\bar{s}_0^i), \pi(\bar{a}_1^i|\bar{s}_1^i), \cdots, \pi(\bar{a}_h^i|\bar{s}_h^i)\}$ in an ascending order, such that $\pi(\bar{a}_j^i|\bar{s}_j^i) \leq \pi(\bar{a}_{j+1}^i|\bar{s}_{j+1}^i), \quad j\in[0,h-1]$
                \STATE Choose $s(\tau_i)=\pi(\bar{a}_{\lfloor\beta h\rfloor}^i|\bar{s}_{\lfloor\beta h\rfloor}^i)$ as the criterion of $\tau_i$.
            \ENDFOR
            \STATE Select $N=\min\{N,|\caD|\}$ trajectories $\{\bar\tau\}_1^N$ with the highest $s(\tau)$ as a new curriculum.
            \STATE Initialize a new replay buffer $\caB$ with $\{\bar\tau\}_1^N$.
            \STATE $\caD = \caD\backslash\{\bar\tau\}_1^N$.
            \FOR{$n=1\to L\times N$}
                \STATE Draw a random batch $\{(s,a)\}_1^B$ from $\caB$.
                \STATE Update $\pi_{\theta}$ using behavior cloning
                $$ \theta\leftarrow\theta- \eta\nabla_\theta\sum_{j=1}^B \left[-\log\pi_\theta(a_j|s_j)\right]$$
            \ENDFOR
            \STATE Update the return filter $V\leftarrow(1-\alpha)V+\alpha\cdot \min\{R(\bar\tau)\}_{1}^N$.
            \STATE Filter $\caD$ by $\caD=\{\tau\in\caD\mid R(\tau)\ge V\}$.
        \ENDWHILE
    \end{algorithmic}
\end{algorithm}
\section{Proofs}
\subsection{Proof for \protect \theo{thm:1}}
\label{ap:proof}

We introduce useful lemmas before providing our proof.
\begin{lemma}[Total variation distance of joint distributions]\label{lm:a1}
Given two joint distributions $\rho_1(x,y)=\rho_1(x|y)\rho_1(y)$ and $\rho_2(x,y)=\rho_2(x|y)\rho_2(y)$, then their total variation distance can be bounded by
\begin{equation*}
    \TV(\rho_1(x,y)\|\rho_2(x,y)) \le \TV(\rho_1(y)\|\rho_2(y)) + \E_{y\sim\rho_2}\left[\TV(\rho_1(x|y)\|\rho_2(x|y))\right]
\end{equation*}
\end{lemma}
\begin{proof}
Directly extend the left hand side, we have
\begin{equation*}\begin{aligned}
    \TV(\rho_1(x,y)\|\rho_2(x,y)) &= \frac12\sum_{x,y} |\rho_1(x,y) - \rho_2(x,y)| \\
    &= \frac12\sum_{x,y} |\rho_1(x|y)\rho_1(y) - \rho_2(x|y)\rho_2(y)| \\
    &= \frac12\sum_{x,y} |\rho_1(x|y)(\rho_1(y) - \rho_2(y)) + (\rho_1(x|y) - \rho_2(x|y))\rho_2(y)|  \\
    &\le \frac12\sum_{x,y}\rho_1(x|y)|(\rho_1(y) - \rho_2(y))| + \frac12\sum_{x,y} |(\rho_1(x|y) - \rho_2(x|y))|\rho_2(y) \\
    &= \frac12\sum_{y}|(\rho_1(y) - \rho_2(y))| + \frac12\sum_{x,y} |(\rho_1(x|y) - \rho_2(x|y))|\rho_2(y) \\
    &= \TV(\rho_1(y)\|\rho_2(y)) + \E_{y\sim\rho_2(y)}[\TV(\rho_1(x|y)\|\rho_2(x|y))]
\end{aligned}\end{equation*}
\end{proof}

Therefore, we have the following proposition.

\begin{proposition}\label{prop:a1}
\begin{equation}\label{eq:propa1}
    \TV(\rho_{\pi}(s,a)\|\rho_{\pi_b}(s,a)) \le \TV(\rho_{\pi}(s)\|\rho_{\pi_b}(s)) + \E_{s\sim\rho_{\pi_b}(s)}\left[\TV(\pi(a|s)\|\pi_b(a|s))\right]
\end{equation}
\end{proposition}
\begin{proof}
Applying \lm{lm:a1} to $\rho_{\pi}(s,a)=\rho_{\pi}(s)\pi(a|s)$ and $\rho_{\pi_b}(s,a)=\rho_{\pi_b}(s)\pi_b(a|s)$ and completes the proof.
\end{proof}

\begin{lemma}[Lemma 5.1 of Xu et. al~\cite{xu2019onvalue}]\label{lm:1}
Let $\Pi$ be the set of all deterministic policy and $|\Pi|=|\caA|^{|\caS|}$. Assume that there does not exist a policy $\pi\in\Pi$ such that $\pi(s^i)=a^i, \forall i\in\{1,\cdots,|\caD|\}$. Then, for any $\delta\in(0,1)$, with probability at least $1-\delta$, the following inequality holds:
\begin{equation}\label{eq:lm2}
    \E_{s\sim\rho_{\pi_b}(s)}[\TV(\pi(a|s)\|\pi_b(a|s))] \le \frac{1}{|\caD|}\sum_{i=1}^{|\caD|} \bbI\left[ \pi(s^i)\neq a^i \right] + \left[\frac{\log|\Pi|+ \log(2/\delta)}{2|\caD|}\right]^{\frac{1}{2}}
\end{equation}
\end{lemma}
% \lm{lm:1} bounds the expectation of total variance between the demonstrated policy $\pi$ and the behavior policy $\hat{\pi}_b$ by two terms, among them the first depends on how much BC learns from the offline dataset, and the second can be seen as the generalization error. 

We are now ready to give the proof for \theo{thm:1}.

\setcounter{theorem}{0}
\begin{theorem}[Performance bound of BC]\label{thm:a1}
Let $\Pi$ be the set of all deterministic policy and $|\Pi|=|A|^{|S|}$. Assume that there does not exist a policy $\pi\in\Pi$ such that $\pi(s^i)=a^i, \forall i\in\{1,\cdots,|\caD|\}$. Let $\hatpi_b$ be the empirical behavior policy and the corresponding state marginal occupancy is $\hatrhob$. Suppose BC begins from initial policy $\pi_0$, and define $\rho_{\pi_0}$ similarly. Then, for any $\delta>0$, with probability at least $1-\delta$, the following inequality holds:
\begin{small}

\begin{equation}\label{eq:thma1}\begin{aligned}
    &\TV(\rho_{\pi}(s,a)\|\rho_{\pi_b}(s,a)) \le \frac{1}{2}\sum_{s\notin\caD}\rho_{\pi_b}(s) + \frac{1}{2}\sum_{s\notin\caD}|\rho_\pi(s)-\rho_{\pi_0}(s)| +  \frac{1}{2}\sum_{s\notin\caD}|\rho_{\pi_0}(s)-\rho_{\pi_b}(s)|  \\
    &+ \frac{1}{2}\sum_{s\in\caD}|\rho_\pi(s)-\hatrhob(s)| + \frac{1}{|\caD|}\sum_{i=1}^{|\caD|} \sI\left[ \pi(s^i)\neq a^i \right] +  \left[\frac{\log|\caS|+\log(2/\delta)}{2|\caD|}\right]^{\frac{1}{2}} + \left[\frac{\log|\Pi|+\log(2/\delta)}{2|\caD|}\right]^{\frac{1}{2}} 
\end{aligned}\end{equation}
\end{small}
\end{theorem}

\begin{proof}
By \lm{lm:1}, for any $\delta\in(0,1)$, with probability at least $1-\delta$, the second term in \eq{eq:propa1} is bounded by:
\begin{equation}\label{eq:pfthm1}
    \E_{s\sim\rho_{\pi_b}(s)}\left[\TV(\pi(a|s)\|\pi_b(a|s))\right] \le \frac{1}{|\caD|}\sum_{i=1}^{|\caD|} \bbI\left[ \pi(s^i)\neq a^i \right] + \left[\frac{\log|\Pi|+\log(2/\delta)}{2|\caD|}\right]^{\frac{1}{2}}
\end{equation}
Then we focus on the first term in \eq{eq:propa1}. Introducing a new distribution $\hatrhob(s)$, the triangle inequality goes that:
\begin{equation}\label{eq:pfthm2}\begin{aligned}
    \TV(\rho_{\pi}(s)\|\rho_{\pi_b}(s)) &\le \TV(\rho_{\pi}(s)\|\hatrhob(s)) + \TV(\hatrhob(s)\|\rho_{\pi_b}(s)) \\
    &= \frac{1}{2}\sum_{s\in\caS}|\rho_\pi(s)-\hatrhob(s)| + \TV(\hatrhob(s)\|\rho_{\pi_b}(s)) \\
    &= \frac{1}{2}\sum_{s\notin\caD}\rho_\pi(s) + \frac{1}{2}\sum_{s\in\caD}|\rho_\pi(s)-\hatrhob(s)| + \TV(\hatrhob(s)\|\rho_{\pi_b}(s)) \\
    &\le \frac{1}{2}\sum_{s\notin\caD}\rho_{\pi_b}(s) + \frac{1}{2}\sum_{s\notin\caD}|\rho_\pi(s)-\rho_{\pi_b}(s)| + \frac{1}{2}\sum_{s\in\caD}|\rho_\pi(s)-\hatrhob(s)| \\
    &+ \TV(\hatrhob(s)\|\rho_{\pi_b}(s)) \\
    &\le \frac{1}{2}\sum_{s\notin\caD}\rho_{\pi_b}(s) + \frac{1}{2}\sum_{s\notin\caD}|\rho_\pi(s)-\rho_{\pi_0}(s)| + \frac{1}{2}\sum_{s\notin\caD}|\rho_{\pi_0}(s)-\rho_{\pi_b}(s)| \\
    &+ \frac{1}{2}\sum_{s\in\caD}|\rho_\pi(s)-\hatrhob(s)| + \TV(\hatrhob(s)\|\rho_{\pi_b}(s)) 
\end{aligned}\end{equation}
Denote $\caS_\caD=\{s\mid s\in\caD\}$. Noticing that $\E_{\caS_\caD\sim\rho_{\pi_b}(s)}[\hatrhob(s)]=\rho_{\pi_b}(s)$, by union bound and Hoeffding's inequality, the following inequality holds:
\begin{equation}\label{eq:pfthm3}\begin{aligned}
    P[\TV(\hatrhob(s)\|\rho_{\pi_b}(s))>\epsilon] &= P[\exists s\in\caS, |\hatrhob(s)-\rho_{\pi_b}(s)|>\epsilon] \\
    &\le \sum_{s\in\caS}P[|\hatrhob(s)-\rho_{\pi_b}(s)|>\epsilon] \\
    &\le 2|\caS|e^{-2|\caD|\epsilon^2}
\end{aligned}\end{equation}
Let $\delta$ be the right hand side, we obtain that with probability at least $1-\delta$, $\TV(\hatrhob(s)\|\rho_{\pi_b}(s))$ is bounded by:
\begin{equation}\label{eq:pfthm4}
    \TV(\hatrhob(s)\|\rho_{\pi_b}(s)) \le \left[\frac{\log|\caS|+\log(2/\delta)}{2|\caD|}\right]^{\frac{1}{2}}
\end{equation}
Combining \prop{prop:a1}, \ineq{eq:pfthm1}, \ineq{eq:pfthm2} and \ineq{eq:pfthm4} completes the proof.
\end{proof}

\subsection{Proof for \protect \observe{obs:find-tau}}
\setcounter{observation}{0}
\label{proof:find-tau}
\begin{observation}\label{obs:b2}
Under the assumption that each trajectory $\tau_{\tilde{\pi}}$ in the dataset $\caD$ is collected by an unknown deterministic behavior policy $\tilde{\pi}$ with an exploration ratio $\beta$. The requirement of the KL divergence constraint $\bbE_{\tilde{\pi}}\left[\kld(\tilde{\pi}(\cdot|s)\|\pi(\cdot|s))\right] \leq \epsilon$ suffices to finding a trajectory that at least $1-\beta$ state-action pairs are sampled by the current policy $\pi$ with a probability of more than $\epsilon_c$ such that $\epsilon_c\geq 1 / \exp{\epsilon}$, \textit{i.e.}:
\begin{equation}\label{eq:b2-1}
    \bbE_{(s,a)\in\tau_{\tilde\pi}}[\sI(\pi(a|s)\geq \epsilon_c)] \geq 1-\beta~,
\end{equation}
% where $\epsilon_c\geq 1 / \exp{\epsilon}$.
\end{observation}
\begin{proof}
We begin with the KL divergence constraint:
\begin{equation}\label{eq:b2-2}
\begin{aligned}
  && \bbE_{\tilde{\pi}}\left[\kld(\tilde{\pi}(\cdot|s)\|\pi(\cdot|s))\right] &\leq \epsilon \\ 
  &\Rightarrow& \bbE_{\tilde{\pi}} \left [\log{\frac{\tilde{\pi}(a|s)}{\pi(a|s)}}\right] &\leq \epsilon \\
  &\Rightarrow& \bbE_{(s,a)\in \tau_{\tilde{\pi}}} \left [\log{\pi(a|s)}\right] &\geq \log{(1-\beta)} - \epsilon\\
  &\Rightarrow& \log{\bbE_{(s,a)\in \tau_{\tilde{\pi}}} \left [\pi(a|s)\right]} &\geq \log{(1-\beta)} - \epsilon\\
  &\Rightarrow& \bbE_{(s,a)\in \tau_{\tilde{\pi}}} \left [\pi(a|s)\right] &\geq \frac{(1-\beta)}{\exp{\epsilon}}
\end{aligned}
\end{equation}

Besides, to achieve $\bbE_{(s,a)\in\tau_{\tilde\pi}}[\sI(\pi(a|s)\geq \epsilon_c)]\geq 1-\beta$, for any state-action pair in $\tau_{\tilde\pi}$, there are at least $1-\beta$ of them can be sampled by $\pi$ with at least the probability of $\epsilon_c$. Therefore, we have the following lower bound:
\begin{equation}\label{eq:b2-3}
    \bbE_{(s,a)\in \tau_{\tilde{\pi}}}[\pi(a|s)] \geq (1-\beta) \cdot \epsilon_c + 0\cdot\beta = (1-\beta) \cdot \epsilon_c
\end{equation}
Combing \ineq{eq:b2-2} and \ineq{eq:b2-3} completes the proof.

\end{proof}

\section{Implementation Details}

\subsection{Implementation for Estimating the Empirical Discrepancy}
\label{ap:implement-dis}

In this part we explain how we estimate the empirical discrepancy outside the dataset, \textit{i.e.}, the term $\frac{1}{2}\sum_{s\notin\caD}|\hat{\rho}_\pi(s)-\hat{\rho}_{\pi_0}(s)|$ in \theo{thm:1}.

\begin{figure}[htbp]
    \centering
    \includegraphics[width=0.575\linewidth]{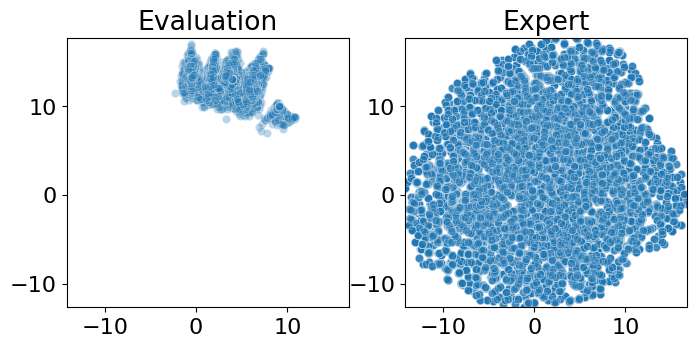}
    \caption{An illustration to the UMAP transformation. Left: states sampled by a random policy in 2-dimensional space transformed by UMAP. Right: expert data to train UMAP in 2-dimensional space.}
    \label{fig:umap}
\end{figure}

Due to the difficulty of estimating the empirical state marginal occupancy measure $\hat{\rho}_\pi(s)$ for a policy $\pi$, especially in continuous or high-dimensional tasks, we estimation its value in 2-dimensional space by projection. 

In detail, we choose UMAP~\cite{mcinnes2018umap} as the projection algorithm and train the projecting function in Hopper using 64\,000 transitions sampled by the expert agent. To evaluate a policy, we sample the same number of transitions, and then project them onto a 2-dimensional space by the trained projecting function. ~\fig{fig:umap} illustrates the projected state distribution, where the left denote the state sampled by a random policy, and the right is the expert data.

For empirical estimation, we subsequently discretize the projected 2-dimensional state space into small grid regions, and estimated the distribution via Kernel Density Estimation (KDE)~\cite{rosenblatt1956remarks} with Gaussian kernel. 
%It is a natural thought that the number of points in a square $\Delta$ can reflect the true probability density at this area.
Suppose $g_{\pi}(\Delta)$ is the Gaussian kernel density function of the transformed data sampled by policy $\pi$, and the dataset is $\caD$. Denote the region that contains state $s$ as $\Delta(s)$. Then the empirical discrepancy can be approximated by
\begin{equation}
    \frac{1}{2}\sum_{s\neq D} |\hat{\rho}_\pi(s)-\hat{\rho}_{\pi_0}(s)| \approx \frac{1}{2}\sum_{\text{all }\Delta}|g_{\pi}(\Delta) - g_{\pi_0}(\Delta))| - \frac{1}{2}\sum_{s\in\caD}|g_\pi(\Delta(s))-g_{\pi_0}(\Delta(s))|
\end{equation}

\subsection{Implementation for the Main Experiment}
The implementation of COIL and BC are based on a Pytorch code framework\footnote{\url{https://github.com/Ericonaldo/ILSwiss}}. As for compared baselines, we take their official implementation:
\begin{itemize}
    \item CQL~\cite{kumar2020conservative}: \url{https://github.com/aviralkumar2907/CQL}
    \item BAIL~\cite{chen2020bail}: \url{https://github.com/lanyavik/BAIL}
    \item AWR~\cite{peng2019advantage}: \url{https://github.com/xbpeng/awr}
\end{itemize}
It should be noted that the public code of AWR is for online RL tasks and we have to modify the code to obtain an offline version, following the instructions in their paper~\cite{peng2019advantage}.

\section{Important Hyperparameters}
\label{se:hyperparameter}

\begin{table*}[htbp]
\caption{Important Hyperparameters.}
\label{tb:hypers}
\small{rd: random\quad mr: medium-replay\quad md: medium\quad me: medium-expert}
\vspace{1ex}
  \centering
  \resizebox{1.0\textwidth}{!}{
    \begin{tabular}{|l|c|c|c|c|c|c|c|c|c|c|c|c|c|c|c|}
    \hline
    \multirow{2}*{Environments} & \multicolumn{5}{c|}{Hopper} & \multicolumn{5}{c|}{Walker2d} & \multicolumn{5}{c|}{HalfCheetah} \\
    \cline{2-16}
    & final & rd & mr & md & me & final & rd & mr & md & me & final & rd & mr & md & me \\
    \hline
    Optimizer & \multicolumn{15}{c|}{AdamOptimizer} \\
    \hline
    Discount factor $\gamma$ & \multicolumn{15}{c|}{0.99} \\
    \hline
    Batch size & \multicolumn{15}{c|}{256} \\
    
    %\hline
    % Filter window size $\alpha$ & \multicolumn{2}{c|}{0.85} & 0.8 & \multicolumn{2}{c|}{0.9} & \multicolumn{4}{c|}{0.85} & 0.9 & 0.8 & \multicolumn{3}{c|}{0.85} & 0.9 \\
    \hline
    Tuning range of the filter window size $\alpha$ & \multicolumn{15}{c|}{[0.8, 0.85, 0.9]} \\
    %\hline
    % Number of selected trajectories $N_0$ & \multicolumn{2}{c|}{1} & \multicolumn{3}{c|}{2} & \multicolumn{2}{c|}{2} & 1 & 2 & 10 & 5 & 2 & \multicolumn{2}{c|}{1} & 10 \\
    \hline
    Tuning range of the number of selected trajectories $N$ & \multicolumn{9}{c|}{[1,2]} & [5,10] & \multicolumn{4}{c|}{[1,2,5]} & [5,10] \\
    %\hline
    % Number of gradient steps for each curriclum $L$ & \multicolumn{3}{c|}{50} & \multicolumn{2}{c|}{100} & 200 & \multicolumn{2}{c|}{50} & \multicolumn{2}{c|}{100} & 400 & 100 & 400 & \multicolumn{2}{c|}{200} \\
    \hline
    Tuning range of $L$ & \multicolumn{10}{c|}{[50, 100, 200]} & \multicolumn{5}{c|}{[100, 200, 400]} \\
    %\hline
    % Pre-train step & \multicolumn{3}{c|}{0} & \multicolumn{2}{c|}{20000} & \multicolumn{3}{c|}{0} & \multicolumn{2}{c|}{40000} & \multicolumn{3}{c|}{0} & \multicolumn{2}{c|}{20000} \\
    %\hline Policy learning rate $\eta_\pi$ & 1e-4 & \multicolumn{3}{c|}{3e-5} & 1e-4 & \multicolumn{2}{c|}{3e-5} & \multicolumn{8}{c|}{1e-4} \\
    \hline 
    Tuning range of $\eta_\pi$ & \multicolumn{15}{c|}{[3e-5, 1e-4]} \\
    \hline
    \end{tabular} 
    }
\end{table*}

The main hyperparameters used in our experiments are shown in \tb{tb:hypers}. Based on the evaluation results of the terminating policy of COIL, might be helpful guidelines for utilizing COIL with different tuning choices are as follows:

\textbf{Policy learning rate $\eta_{\pi}$.} Similar to CQL~\cite{kumar2020conservative}, we evaluated COIL with a policy learning rate in the range of $[3e-5, 1e-4]$. We find that $1e-4$ almost uniformly attain good performance and we chose $1e-4$ as the default across all datasets. Besides, we recommend increasing the number of gradient steps $L$ to be compatible with a low learning rate on the same task.

\textbf{Gradient steps for each curriculum $L$.} The gradient steps correspond to how long that BC should be utilized for imitating the target policy at the current level. For easier tasks as Hopper and Walker, we evaluate COIL in the range of $[50, 100, 200]$ and for harder tasks like Halfcheetah, we tune in the range of $[100, 200, 400]$. The default choice is 100. 

\textbf{Number of selected trajectories $N$ and Filter window size $\alpha$.} These two hyperparameters affect the experimental results more significantly. Moreover, as mentioned in \se{exp:ablation}, they can be tuned based on the distribution of the dataset. Detailed guidelines can be found in \se{exp:ablation} and \ap{ap:ablation}.

\section{Additional Results}
\label{ap:addition-exps}

\subsection{Results of the Motivating Example}
\label{ap:motivation-results}

Here we provide additional results for the motivated example experiments performed on the Hopper environment. 

% \minghuan{Change this form.}
\begin{table}[hbtp]
\caption{Numerical values of results presented in \fig{fig:motivation-suba}, the means and standard deviations are calculated over 3 random seeds.}
\vspace{6pt}
\centering
\resizebox{0.85\columnwidth}{!}{
\begin{tabular}{c|ccccc}
\toprule
Agent                & Initial         & 1 Trajectory     & 4 Trajectories   & 256 Trajectories & 1024 Trajectories \\
\midrule
\texttt{Random}      & 49.9 (3.0)    & 255.0 (355.1)  & 374.1 (392.8)  & 2802.9 (846.4) & 3004.7 (656.4) \\
\texttt{1/3 Return} & 1362.6 (853.7) & 2930.3 (886.9) & 2709.5 (974.2) & 3096.4 (630.9) & 3310.2 (442.0) \\
\texttt{1/3 Trained} & 3289.1 (3.43) & 3014.4 (926.0) & 3428.4 (392.9) & 3582.2 (115.9) & 3588.3 (163.7) \\
\texttt{2/3 Trained} & 3558.0 (6.0)  & 3609.8 (14.2)  & 3602.6 (80.3)  & 3612.4 (78.8)  & 3610.6 (13.9) \\
\midrule
Demonstration        & 3622.3 (22.0) & -                & -                & -                & - \\
\bottomrule
\end{tabular}
}
\vspace{-8pt}
\label{tab:motivation}
\end{table}

\begin{table}[hbtp]
\caption{Numerical values of results presented in \fig{fig:motivation-subb}, the means and standard deviations are calculated over 3 random seeds.}
\vspace{6pt}
\centering
\resizebox{0.65\columnwidth}{!}{
\begin{tabular}{c|ccc}
\toprule
Agent                & 1 Trajectory    & 4 Trajectories  & 256 Trajectories    \\
\midrule
\texttt{Random}      & 0.706 (0.111) & 0.737 (0.124) & 0.518 (0.077)     \\
\texttt{1/3 Trained} & 0.528 (0.006) & 0.549 (0.007) & 0.340 (0.004)   \\
\texttt{2/3 Trained} & 0.541 (0.001) & 0.551 (0.006) & 0.313 (0.005)      \\
\bottomrule
\end{tabular}
}
\vspace{-8pt}
\label{tab:motivation2}
\end{table}

We do not evaluate the empirical discrepancy in the case of 1024 trajectories because the projected points are so dense that the grid size is required to be rather small, which leads to an unacceptable computing time.

\subsection{Additional Ablation Studies}
\label{ap:ablation}

\subsubsection{Ablation on Trajectory Number \texorpdfstring{$N$}{N}}
\begin{figure}[htbp]
\begin{subfigure}[b]{0.48\textwidth}
\centering
\includegraphics[height=0.48\linewidth]{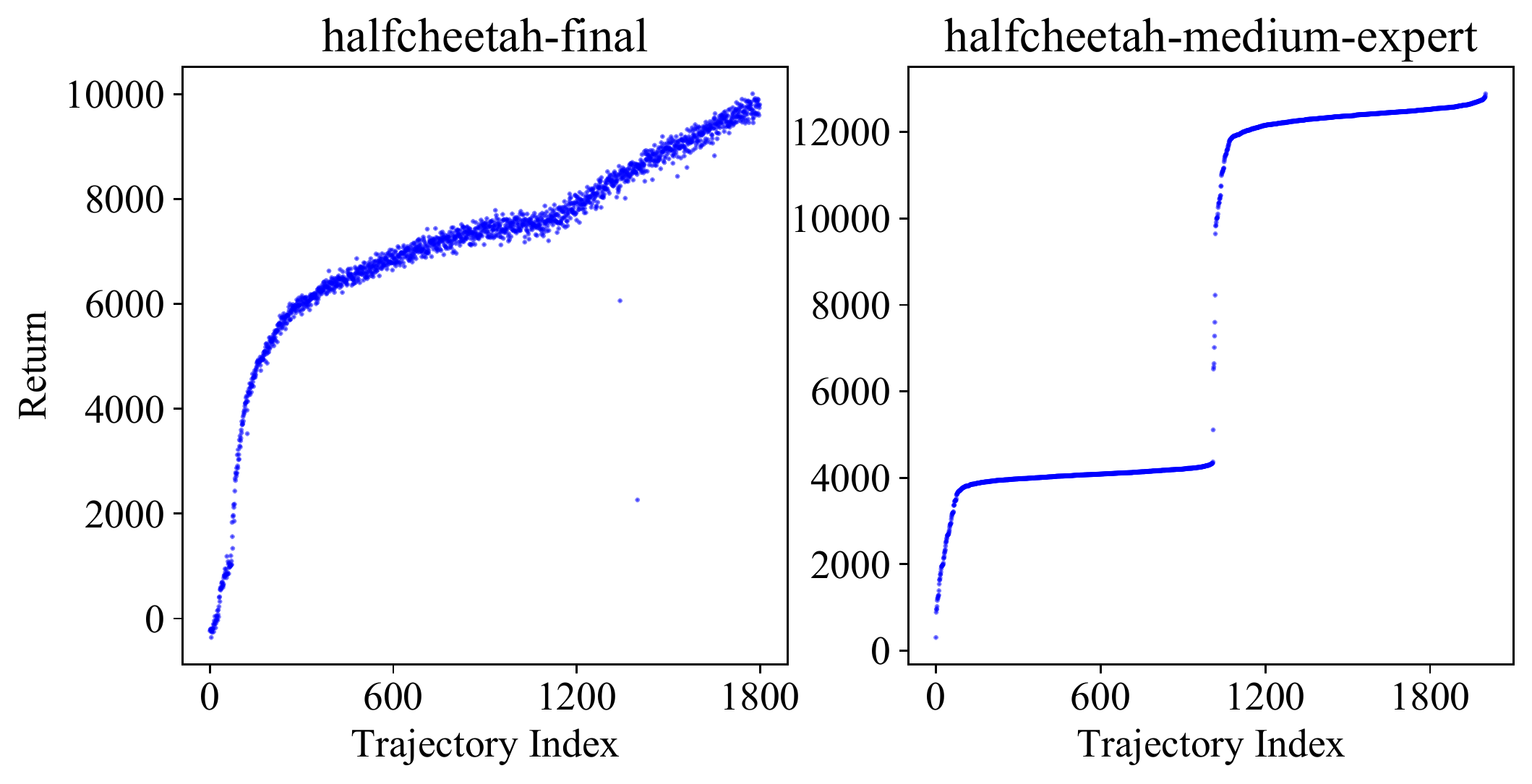}
\caption{Returns of used datasets.}
\label{fig:hc-final-me}
\end{subfigure}
\begin{subfigure}[b]{0.48\textwidth}
\centering
\includegraphics[height=0.45\linewidth]{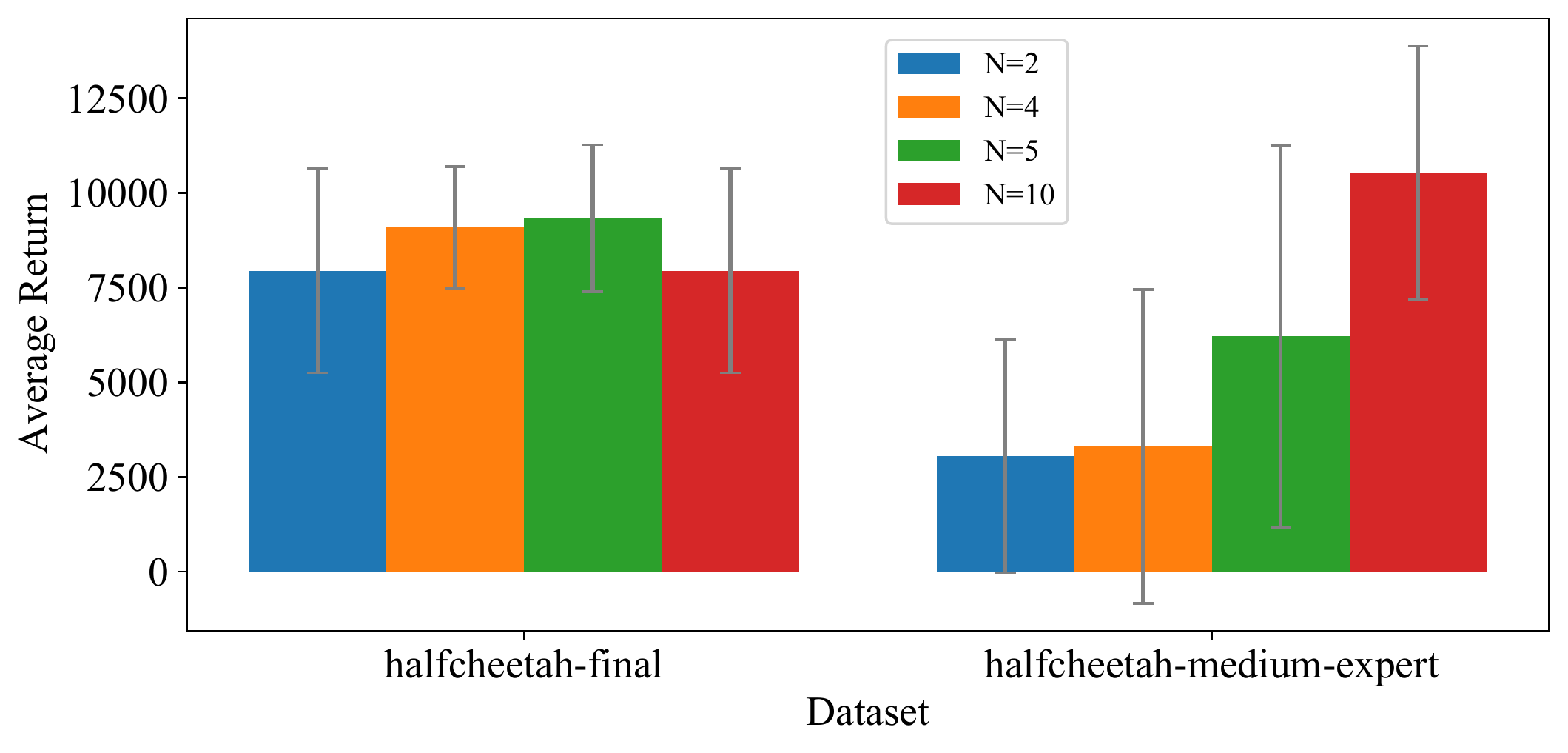}
\caption{Performance with different $N$s.}
\label{fig:ablation-n0}
\end{subfigure}

\caption{Returns of trajectories in \hytt{halfcheetah-final} and \hytt{halfcheetah-medium-expert}, and final performances of COIL on them with different $N$.}
\end{figure}

We further show how the number of chosen trajectories $N$ can be determined by the dataset. \fig{fig:ablation-n0} shows the results on \hytt{halfcheetah-final} and \hytt{halfcheetah-medium-expert} for examples. Obviously,  trajectories in \hytt{halfcheetah-final} are densely and smoothly arranged than those in \hytt{halfcheetah-medium-expert}, indicating that the discrepancy between the behavior policies contained in the final dataset may be smaller. As revealed in \theo{thm:1}, as the distance between the behavior policies becomes farther, more training samples are required for a good imitation. Therefore, a larger $N$ should be chosen for \hytt{halfcheetah-medium-expert} than the other one. \fig{fig:ablation-n0} shows consistent results to our expectation, where a large value of  $N=10$ puts the best performance on \hytt{halfcheetah-medium-expert}, and medium values of $N$ (4 or 5) provide better behaviors on \hytt{halfcheetah-final}.

\subsubsection{Further Ablation on Return Filter}

\begin{figure}[htbp]
\centering
\begin{minipage}{1.0\linewidth}
\vspace{-9pt}
\begin{subfigure}[b]{0.3\textwidth}
\centering
\includegraphics[width=0.88\linewidth,height=0.85\linewidth]{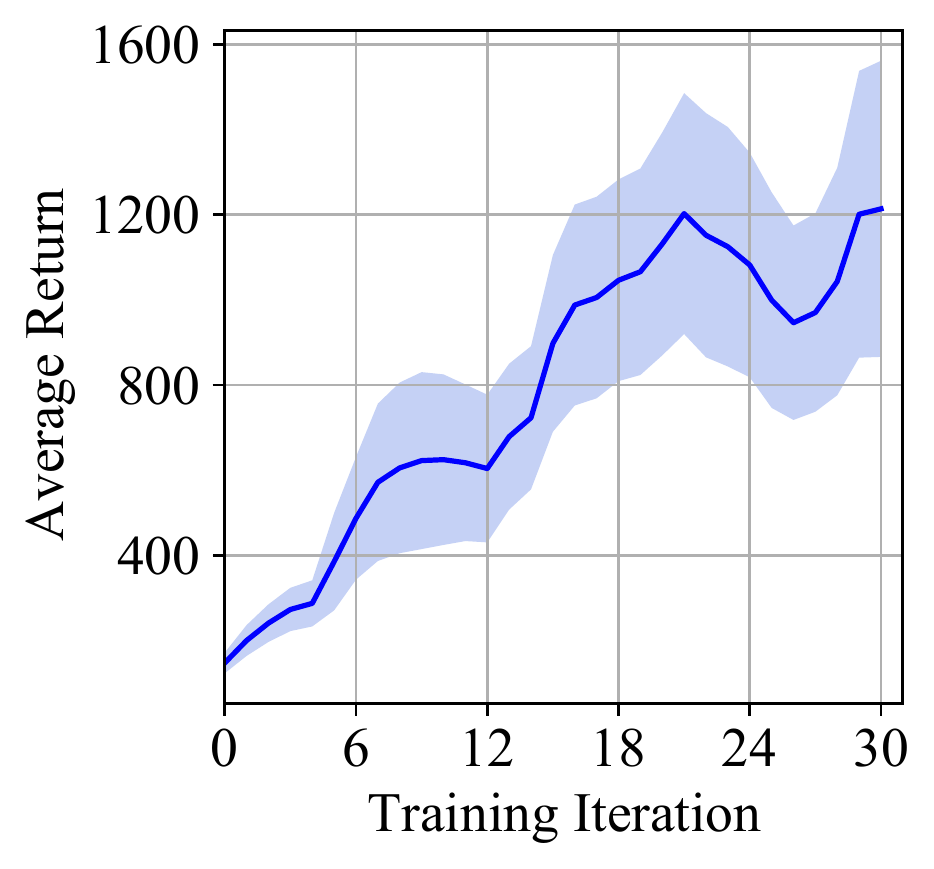}

\vspace{-5pt}
%\caption{Hopper-final.}
\label{fig:hopper-final-return-ft0.0}
\end{subfigure}
\begin{subfigure}[b]{0.3\textwidth}
\centering
\includegraphics[width=0.88\linewidth,height=0.85\linewidth]{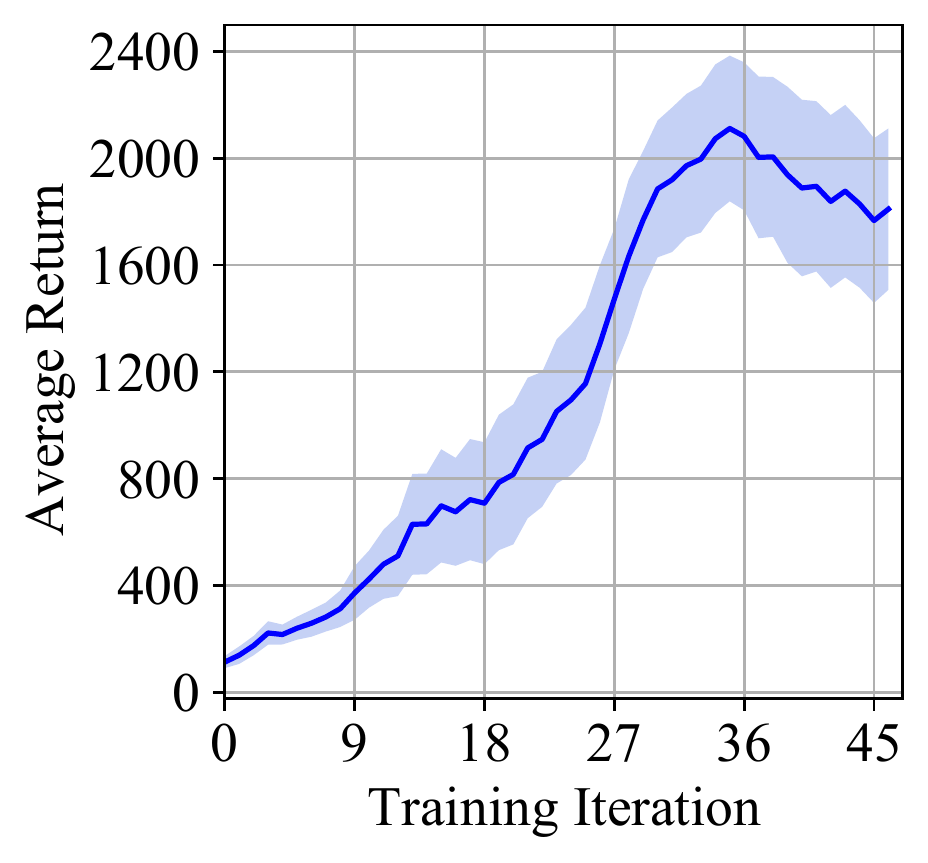}

\vspace{-5pt} 
%\caption{Walker2d-final.}
\label{fig:hopper-final-return-ft0.5}
\end{subfigure}
\begin{subfigure}[b]{0.3\textwidth}
\centering
\includegraphics[width=0.88\linewidth,height=0.85\linewidth]{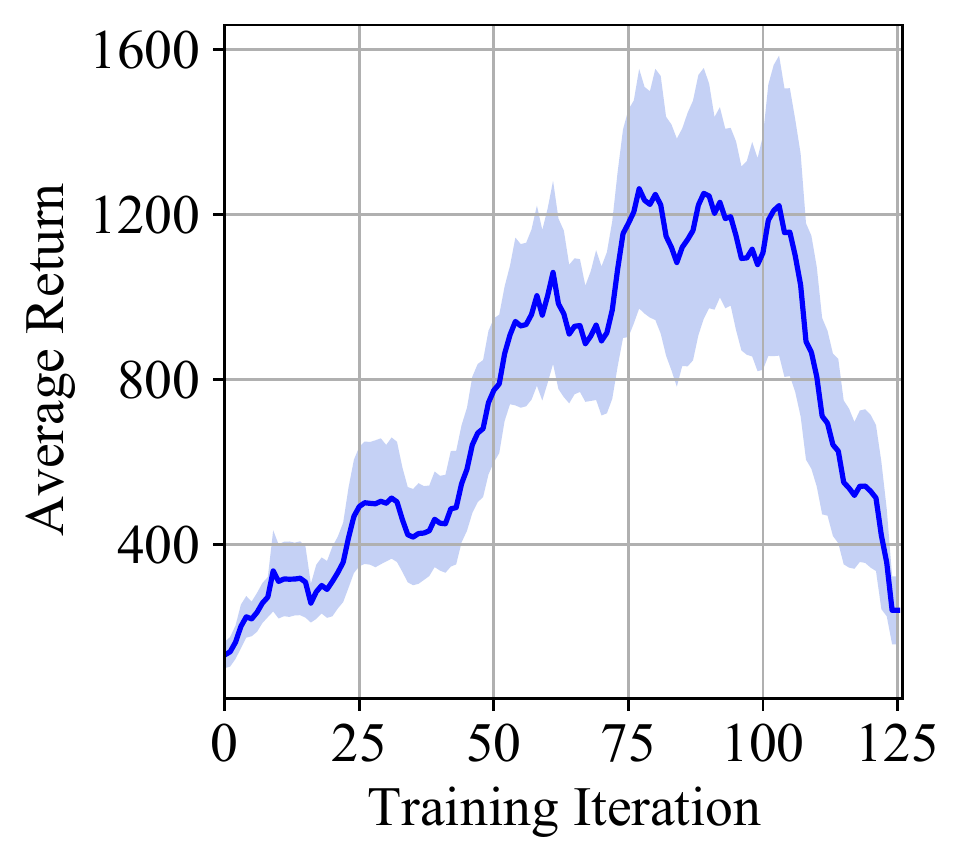}

\vspace{-5pt} 
%\caption{HalfCheetah-final.}
\label{fig:hopper-final-return-ft1.0}
\end{subfigure}

\vspace{2pt}
\begin{subfigure}[b]{0.3\textwidth}
\centering
\includegraphics[width=0.85\linewidth,height=0.85\linewidth]{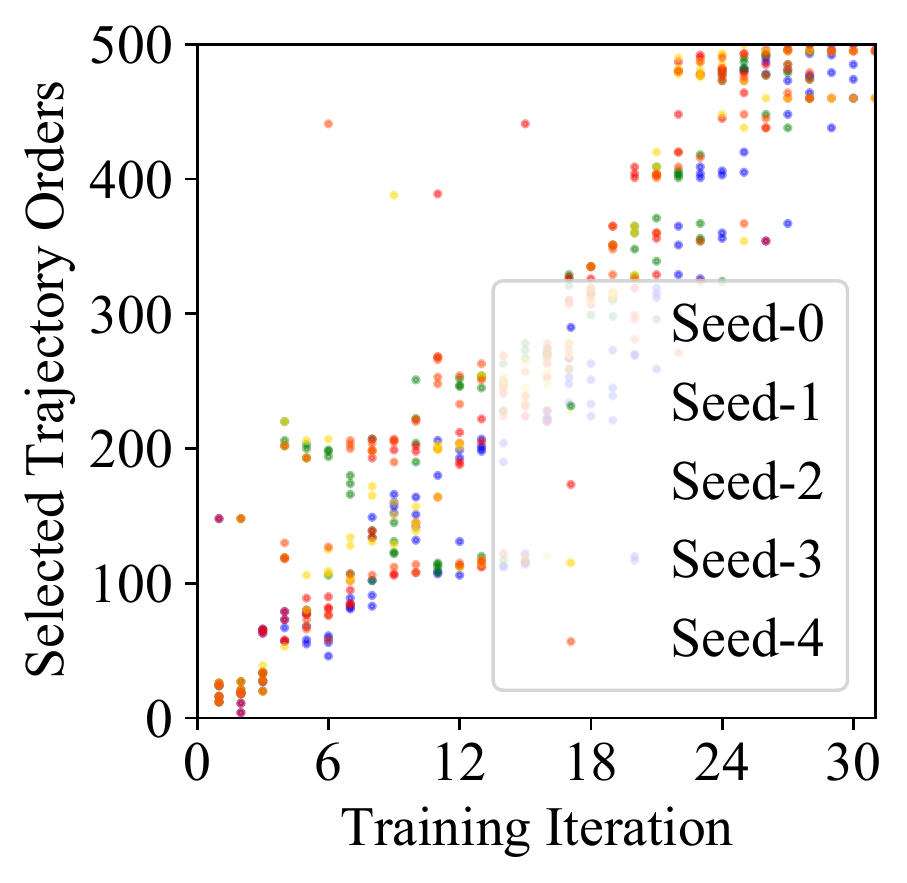}

\vspace{-2pt}
\caption{$\alpha=0.0$.}
\label{fig:hopper-final-traj-ft0.0}
\end{subfigure}
\begin{subfigure}[b]{0.3\textwidth}
\centering
\includegraphics[width=0.85\linewidth,height=0.85\linewidth]{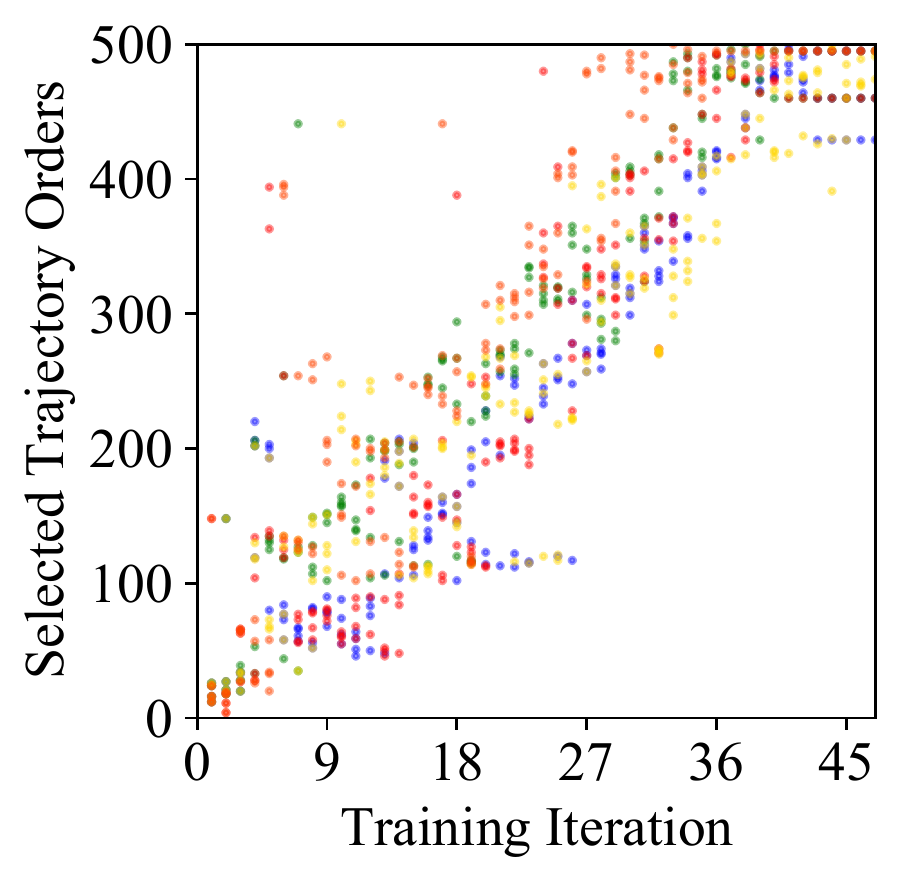}

\vspace{-2pt} 
\caption{$\alpha=0.5$.}
\label{fig:hopper-final-traj-ft0.5}
\end{subfigure}
\begin{subfigure}[b]{0.3\textwidth}
\centering
\includegraphics[width=0.86\linewidth,height=0.85\linewidth]{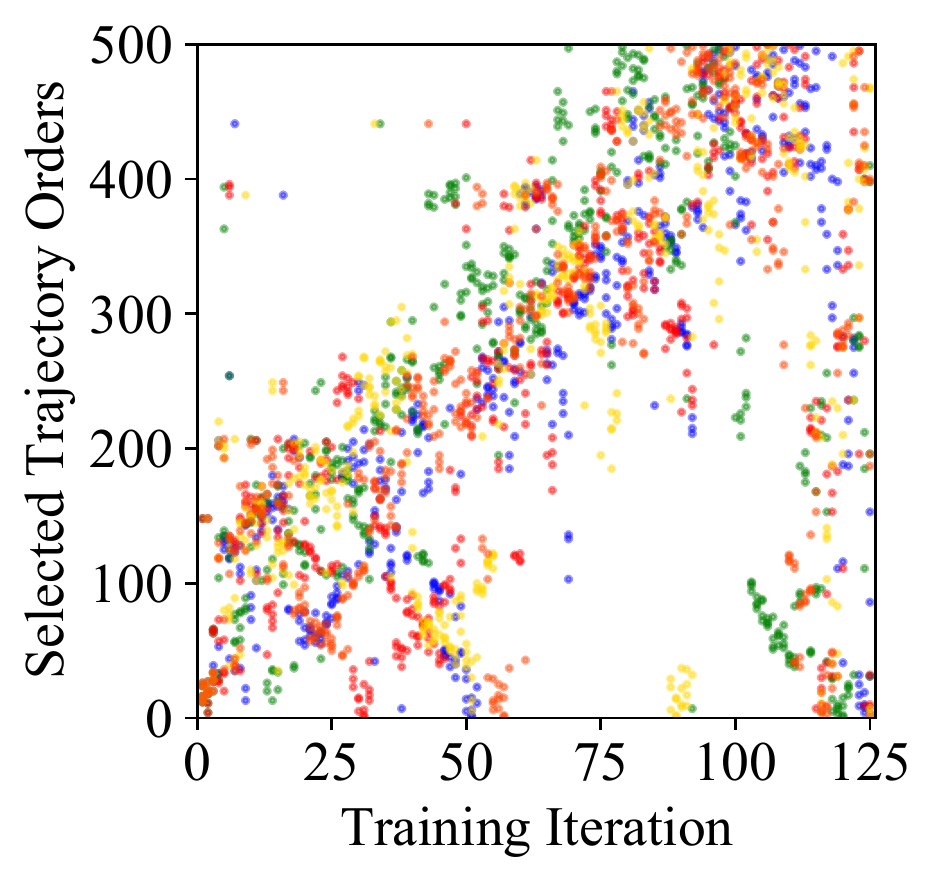}

\vspace{-2pt} 
\caption{$\alpha=1.0$.}
\label{fig:hopper-final-traj-ft1.0}
\end{subfigure}
\end{minipage}

\caption{Training curves and orders of selected trajectories with different $\alpha$ on \texttt{hopper-final}.}
\label{fig:ablation-filter}
\end{figure}

To further illustrate the functionality of the return filter we conduct more ablation experiments on the hyperparameter $\alpha$, where we set $\alpha$ as 1.0 (no return filter), 0.0 (no moving average) and 0.5 (rapid moving average). Obviously, without a return filter ($\alpha$=1.0), the agent imitates earlier trajectories in the final which deteriorates the final performance; without a moving average ($\alpha$=1.0), the agent quickly drops the candidate trajectories which leads it to learn nothing; the results with a rapid moving average ($\alpha$=0.5), are better and more stable but it still fails to imitate the best behavior data. Therefore, the return filter is a key ingredient for COIL that should be designed carefully when applied on different datasets.

\subsection{Complete Training Curves on Final Datasets}
\begin{figure}[tbhp]
\begin{subfigure}[b]{0.9\textwidth}
\centering
\includegraphics[height=0.38\linewidth]{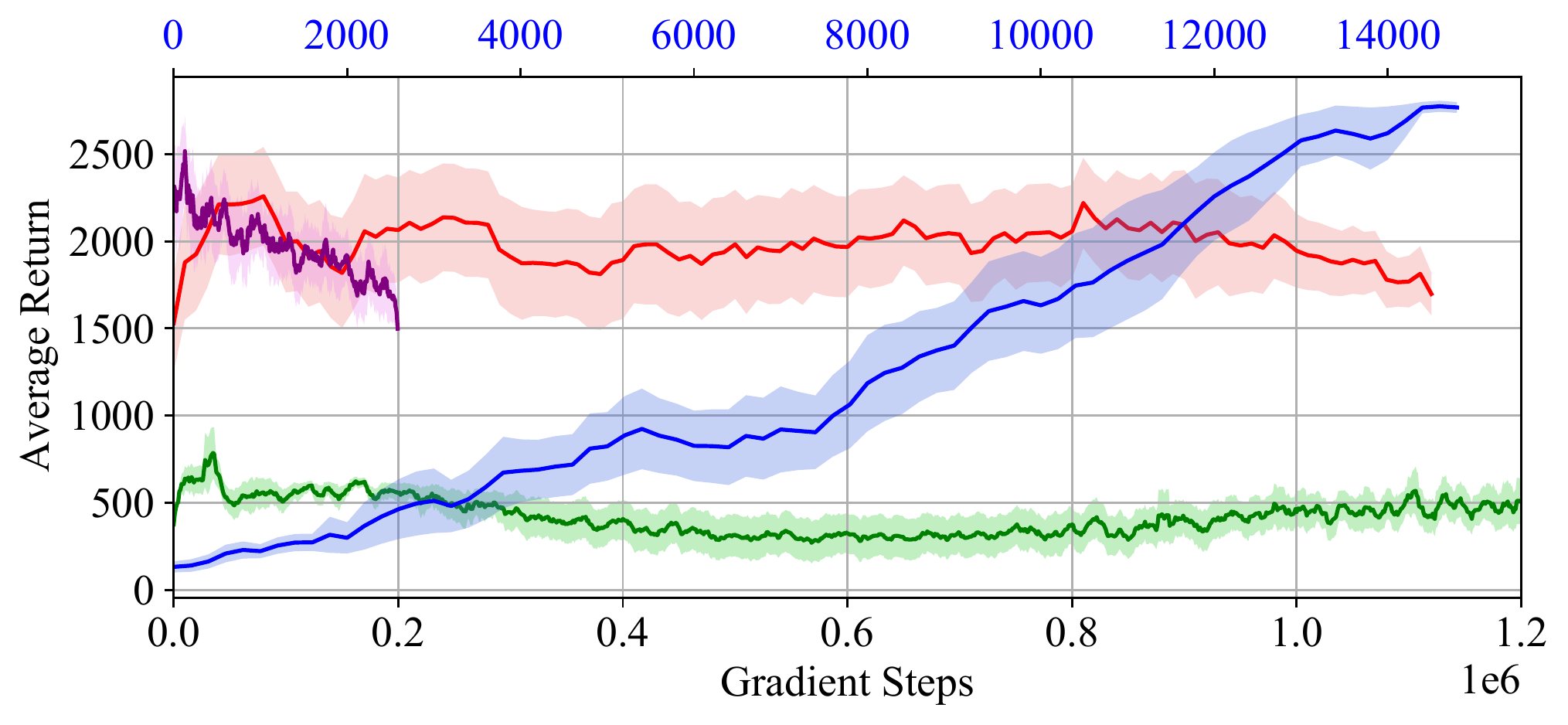}
\caption{Hopper-final.}
\label{fig:hopper-all}
\end{subfigure}

\begin{subfigure}[b]{0.9\textwidth}
\centering
\includegraphics[height=0.38\linewidth]{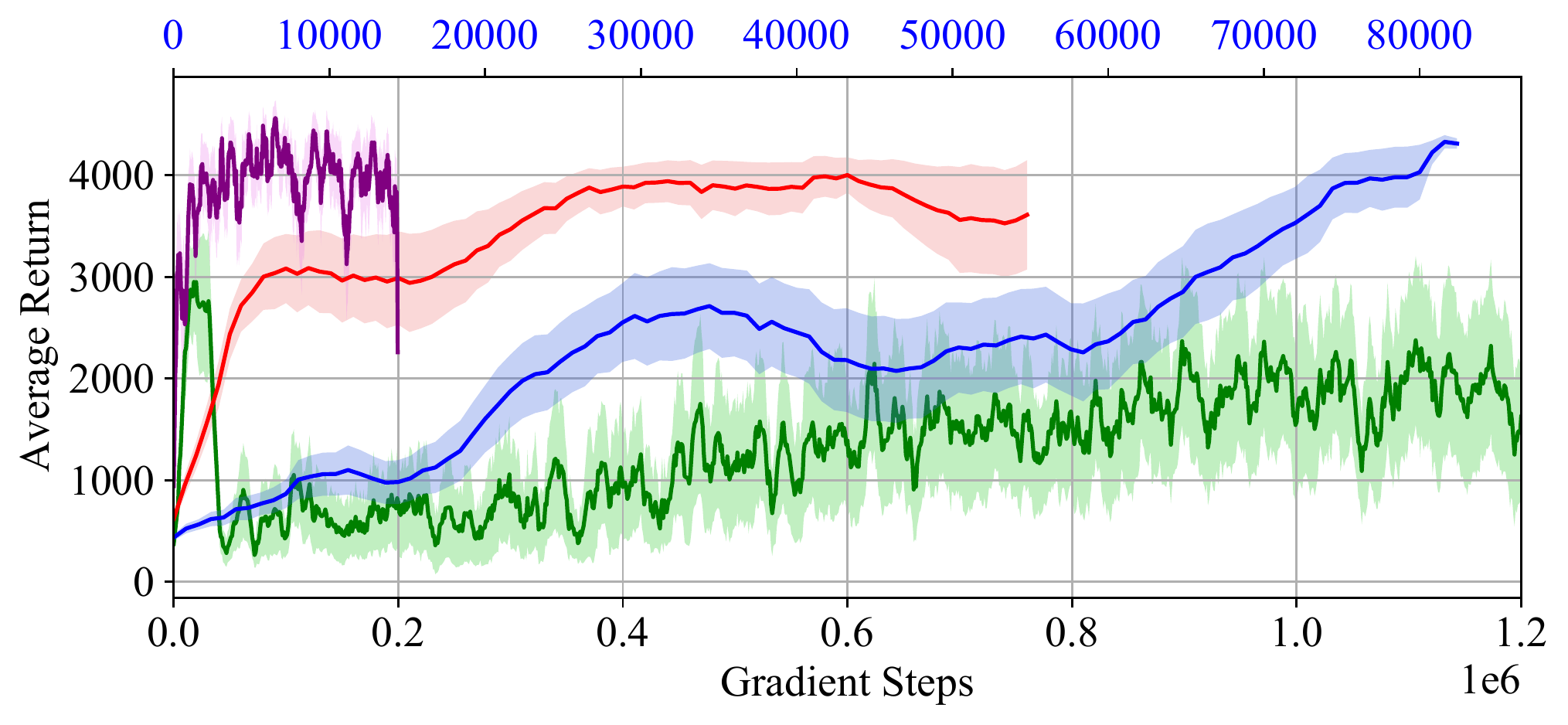}
\caption{Walker2d-final.}
\label{fig:walker2d-all}
\end{subfigure}

\begin{subfigure}[b]{0.9\textwidth}
\centering
\includegraphics[height=0.38\linewidth]{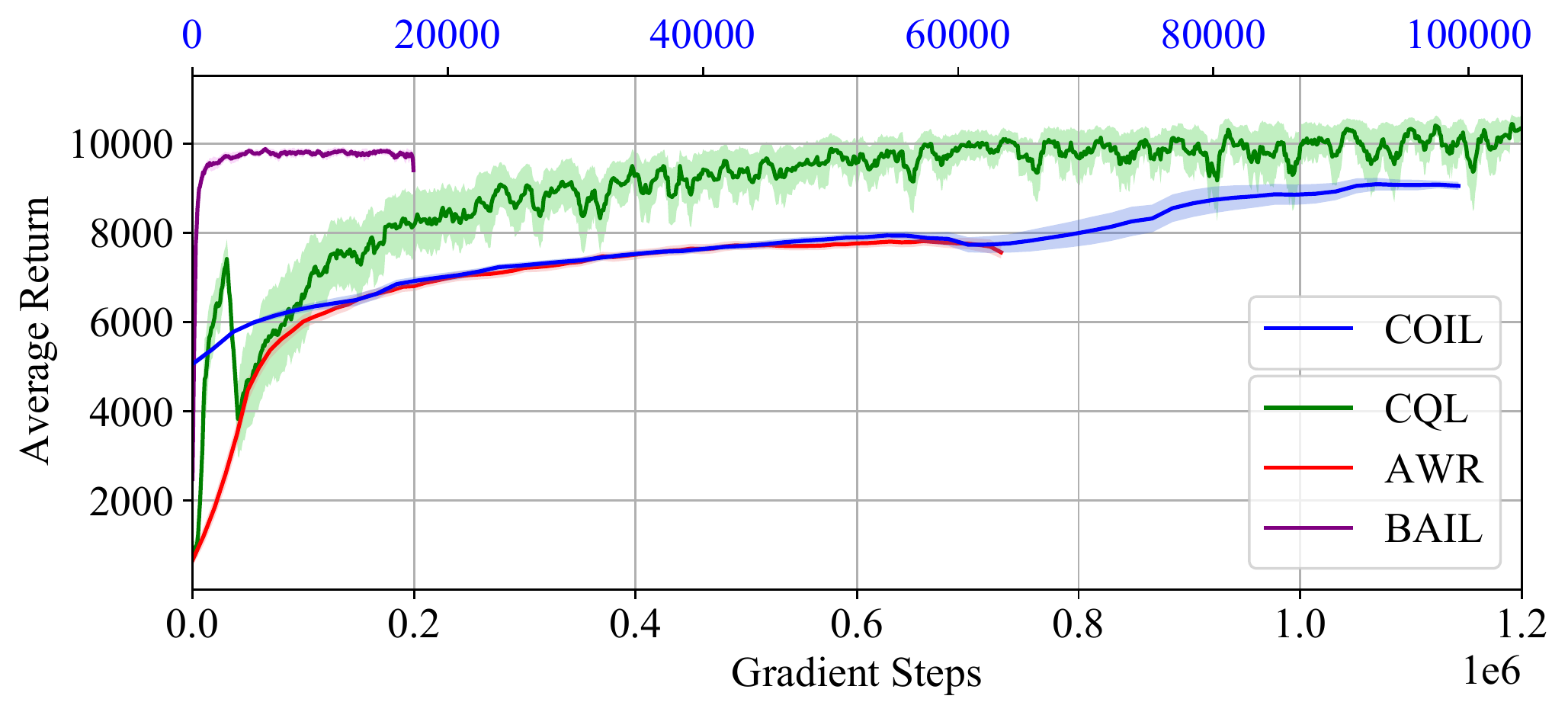}
\caption{HalfCheetah-final.}
\label{fig:halfcheetah-all}
\end{subfigure}

\caption{Comparison of training curves between COIL, CQL, AWR, and BAIL on \textit{final} datasets. Except BAIL has a large batch size (1000), the other methods keep the same batch size (256). Different methods terminate with different gradient steps. The top axis ({\color{blue}{blue}}) of each figure illustrates the gradient step of COIL in a small magnitude, showing the highest efficiency of our method. And the bottom axis ({\color{black}{black}}) denotes the gradient step for the other baselines.}
\label{fig:curve-all}
\end{figure}

We show the complete training curves of COIL, CQL, AWR and BAIL on \textit{final} datasets. We do not cover D4RL benchmarks since those numerical results of baselines are directly borrowed from Fu et al.~\cite{fu2020d4rl}. As is observed in \fig{fig:curve-all}, CQL works well on Halfcheetah; but on Hopper and Walker, the other imitation-based methods are more effective to reach a better performance. 
Saliently, COIL only needs fewer gradient steps to terminate with an excellent policy, such that we have to use a different scale of axis (the top {\color{blue}{axis}}) to illustrate COIL clearly.

\end{document}